\documentclass{article}

\PassOptionsToPackage{numbers, compress}{natbib}

\usepackage[final]{neurips_2019}
\usepackage[utf8]{inputenc} 
\usepackage[T1]{fontenc}    
\usepackage{hyperref}       
\usepackage{url}            
\usepackage{booktabs}       
\usepackage{amsfonts}       
\usepackage{nicefrac}       
\usepackage{microtype}      

\usepackage{booktabs} 

\newcommand\numberthis{\addtocounter{equation}{1}\tag{\theequation}}  
\newcommand{\ignore}[1]{}

\usepackage{multirow}
\usepackage{array}

\usepackage{subcaption}

\usepackage{graphicx} 
\usepackage{tikz}
\usetikzlibrary{arrows.meta,calc,decorations.markings,math,arrows.meta}

\usepackage{amsmath, amssymb, amsthm}
\usepackage{mathtools}
\DeclareMathOperator*{\minimize}{minimize}

\theoremstyle{definition}
\newtheorem{definition}{Definition}[section]
\newtheorem{theorem}{Theorem}[section]
\newtheorem{proposition}{Proposition}[section]

\newtheorem*{theorem*}{Theorem}
\newtheorem*{proposition*}{Proposition}

\usepackage{algorithm,algorithmic}

\newcommand{\cov}{{\Sigma}}
\newcommand{\la}{\mathbb{E}\left[}
\newcommand{\ra}{\right]}
\newcommand{\snr}{\mathit{s}}
\newcommand{\half}{\nicefrac[]{1}{2}}

\title{Fast structure learning with modular regularization}

\author{%
  Greg Ver Steeg \\
  Information Sciences Institute \\
  University of Southern California \\
  Marina del Rey, CA 90292 \\
  \texttt{gregv@isi.edu} \\
  \And
  Hrayr Harutunyan \\
  Information Sciences Institute \\
  University of Southern California \\
  Marina del Rey, CA 90292 \\
  \texttt{hrayrh@isi.edu} \\
  \And
  Daniel Moyer \\
  Information Sciences Institute \\
  University of Southern California \\
  Marina del Rey, CA 90292 \\
  \texttt{moyerd@usc.edu} \\
  \And
  Aram Galstyan \\
  Information Sciences Institute \\
  University of Southern California \\
  Marina del Rey, CA 90292 \\
  \texttt{galstyan@isi.edu}
}

\begin{document}

\maketitle

\begin{abstract}
Estimating graphical model structure from high-dimensional and undersampled data is a fundamental problem in many scientific fields.
Existing approaches, such as GLASSO, latent variable GLASSO, and latent tree models, suffer from high computational complexity and may impose unrealistic sparsity priors in some cases.
We introduce a novel method that leverages a newly discovered connection between information-theoretic measures and structured latent factor models to derive an optimization objective which encourages {\em modular} structures where each observed variable has a single latent parent.
The proposed method has linear stepwise computational complexity w.r.t. the number of observed variables.
Our experiments on synthetic data demonstrate that our approach is the only method that recovers modular structure better as the dimensionality increases. We also use our approach for estimating covariance structure for a number of real-world datasets and show that it consistently outperforms state-of-the-art estimators at a fraction of the computational cost. Finally, we apply the proposed method to high-resolution fMRI data (with more than $10^5$ voxels) and show that it is capable of extracting meaningful patterns.
%

\end{abstract}

\section{Introduction}
\label{sec:intro}
The ability to recover the true relationships among many variables directly from data is a holy grail in many scientific domains, including neuroscience, computational biology, and finance. 
Unfortunately, the problem is challenging in high-dimensional and undersampled regimes due to the curse of dimensionality. Existing methods try to address the challenge by making certain assumptions about the structure of the solution. 
For instance, graphical LASSO, or GLASSO~\cite{friedman2008sparse}, imposes sparsity constraints on the inverse covariance matrix. While GLASSO perfroms well for certain undersampled problems, its computational complexity is cubic in the number of variables, making it impractical for even moderately sized problems.
One can improve the scalability by imposing even stronger sparsity constraints, but this approach fails for many real-world datasets that do not have ultra-sparse structure.
Other methods such as latent variable graphical LASSO (LVGLASSO)~\cite{chandrasekaran2010latent} and latent tree modeling methods~\cite{zhang2017latent} suffer from high computational complexity as well, whereas approaches like PCA, ICA, or factor analysis have better time complexity but perform very poorly in undersampled regimes.



In this work we introduce a novel latent factor modeling approach for estimating multivariate Gaussian distributions.
The proposed method -- linear Correlation Explanation or linear CorEx -- searches for independent latent factors that explain all correlations between observed variables, while also biasing the model selection towards modular latent factor models -- directed latent factor graphical models where each observed variable has a single latent variable as its only parent.
Biasing towards modular latent factor models corresponds to preferring models for which the covariance matrix of observed variables is block-diagonal with each block being a diagonal plus rank-one matrix.
This modular inductive prior is appropriate for many real-world datasets, such as stock market, magnetic resonance imaging, and gene expression data, where one expects that variables can be divided into clusters, with each cluster begin governed by a few latent factors and latent factors of different clusters being close to be independent.
Additionally, modular latent factors are easy to interpret and are popular for exploratory analysis in social science and biology~\cite{cattell}. 
Furthermore, we provide evidence that learning the graphical structure of modular latent factor models with fixed number of latent factors gets \emph{easier} as the number of observed variables increases -- an effect which we call blessing of dimensionality.

We derive the method by noticing that certain classes of graphical models correspond to global optima of information-theoretic functionals. 
The information-theoretic optimization objective for learning unconstrained latent factor models is shown in  Fig.~\ref{fig:schematic-a}. 
We add an extra regularization term that encourages the learned model to have modular latent factors (shown in Fig.~\ref{fig:schematic-b}).
The resulting objective is trained using gradient descent, each iteration of which has \emph{linear} time and memory complexity in the number of observed variables $p$, assuming the number of latent factors is constant.

We conduct experiments on synthetic data and demonstrate that the proposed method is the only one that exhibits a blessing of dimensionality when data comes from a modular (or approximately modular) latent factor model.
Based on extensive evaluations on synthetic as well as over fifty real-world datasets, we observe that our approach handily outperforms other methods in covariance estimation, with the largest margins on high dimensional, undersampled datasets. 
Finally, we demonstrate the scalability of linear CorEx by applying it to high-resolution fMRI data (with more than 100K voxels), and show that the method finds interpretable structures.

\section{Learning structured models}
\label{sec:preliminaries}
\begin{figure}[t]
\begin{subfigure}[t]{0.5\textwidth}
\centering
\scalebox{0.75}{
\begin{tikzpicture}
    \draw (-1.5,1.8) circle (0.5cm) node {$Z_1$};
    \draw (+1.5,1.8) circle (0.5cm) node {$Z_m$};
    \draw (0, 1.8) node {$\ldots$};
    \draw[fill=gray!20] (-2.25,0) circle (0.5cm) node {$X_1$};
    \draw[fill=gray!20] (-0.75,0) circle (0.5cm) node {$X_2$};
    \draw[fill=gray!20] (+0.75,0) circle (0.5cm) node {$X_{\ldots}$};
    \draw[fill=gray!20] (+2.25,0) circle (0.5cm) node {$X_p$};
    \draw[-{Latex[scale=1.5]}] (-1.7,1.34) -- (-2.3,0.5);
    \draw[-{Latex[scale=1.5]}] (-1.4,1.31) -- (-0.85,0.49);
    \draw[-{Latex[scale=1.5]}] (-1.3,1.34) -- (+0.7,0.5);
    \draw[-{Latex[scale=1.5]}] (-1.2,1.4) -- (+2.1,0.476);
    
    \draw[-{Latex[scale=1.5]}] (+1.2,1.4) -- (-2.1,0.476);
    \draw[-{Latex[scale=1.5]}] (+1.3,1.34) -- (-0.7,0.5);
    \draw[-{Latex[scale=1.5]}] (1.4,1.31) -- (+0.85,0.49);
    \draw[-{Latex[scale=1.5]}] (1.7,1.34) -- (+2.25,0.5);
    \draw (0, -0.8) node {$\Updownarrow$};
    \normalsize
    \draw (0, -1.4) node {$TC(X \mid Z) + TC(Z) = 0$};
    \end{tikzpicture}
}
\caption{Unconstrained latent factor model}
\label{fig:schematic-a}
\end{subfigure}%
\hspace{-1cm}
\begin{subfigure}[t]{0.5\textwidth}
\centering
\scalebox{0.75}{
\begin{tikzpicture}
    \draw (-1.5,1.8) circle (0.5cm) node {$Z_1$};
    \draw (+1.5,1.8) circle (0.5cm) node {$Z_m$};
    \draw (0, 1.8) node {$\ldots$};
    \draw[fill=gray!20] (-2.25,0) circle (0.5cm) node {$X_1$};
    \draw[fill=gray!20] (-0.75,0) circle (0.5cm) node {$X_2$};
    \draw[fill=gray!20] (+0.75,0) circle (0.5cm) node {$X_{\ldots}$};
    \draw[fill=gray!20] (+2.25,0) circle (0.5cm) node {$X_p$};
    \draw[-{Latex[scale=1.5]}] (-1.7,1.34) -- (-2.3,0.5);
    \draw[-{Latex[scale=1.5]}] (-1.4,1.31) -- (-0.85,0.49);
    
    \draw[-{Latex[scale=1.5]}] (1.4,1.31) -- (+0.85,0.49);
    \draw[-{Latex[scale=1.5]}] (1.7,1.34) -- (+2.25,0.5);
    \draw (0, -0.8) node {$\Downarrow$ (for any distribution) \hspace{0.2cm} $\Uparrow$ (for Gaussians)};
    \normalsize
    \draw (0, -1.4) node {$TC(X \mid Z) + TC(Z) = 0, \ \& \ \forall i, TC(Z \mid X_i) = 0$};
    \end{tikzpicture}
}
\caption{Modular latent factor model}
\label{fig:schematic-b}
\end{subfigure}%
\caption{Unconstrained and modular latent factor models. Both models admit equivalent information-theoretic characterization (see Prop.~\ref{thm1} and Thm.~\ref{thm2} respectively).}
\label{fig:schematic}
\end{figure}

\textbf{Notation \quad} Let $X \equiv X_{1:p} \equiv (X_1, X_2, \ldots, X_p)$ denote a vector of $p$ observed variables, and let $Z \equiv Z_{1:m} \equiv (Z_1, Z_2, \ldots, Z_m)$ denote a vector of $m$ latent variables. Instances of $X$ and $Z$ are denoted in lowercase, with $x=(x_1, \ldots, x_p)$ and $z=(z_1, \ldots, z_m)$ respectively.
Throughout the paper we refer to several information-theoretic concepts, such as differential entropy: $H(X) = -\mathbb{E}{\log p(x)}$, mutual information: $I(X; Y) = H(X) + H(Y) - H(X,Y)$, multivariate mutual information, historically called total correlation~\cite{watanabe}: $TC(X) = \sum_{i=1}^p H(X_i) - H(X)$, and their conditional variants, such as $H(X|Z) = \mathbb{E}_z\left[H(X|Z=z)\right], TC(X|Z) =\mathbb{E}_z\left[TC(X|Z=z)\right]$. Please refer to \citet{cover} for more information on these quantities.

Consider the latent factor model shown in Fig.~\ref{fig:schematic-a}, which we call unconstrained latent factor model.
In such models, the latent factors explain dependencies present in $X$, since $X_1, \ldots, X_p$ are conditionally independent given $Z$.
Thus, learning such graphical models gives us meaningful latent factors.
Typically, to learn such a graphical model we would parameterize the space of models with the desired form and then try to maximize the likelihood of the data under the model. An alternative way, the one that we use in this paper, is to notice that some types of directed graphical models can be expressed succinctly in terms of information-theoretic constraints on the joint density function. In particular, the following proposition provides an information-theoretic characterization of the unconstrained latent factor model shown in Fig.~\ref{fig:schematic-a}.

\begin{proposition}
\label{thm1}
The random variables $X$ and $Z$ are described by a directed graphical model where the parents of $X$ are in $Z$ and the $Z$'s are independent if and only if $TC(X|Z) + TC(Z)=0$. 
\end{proposition}
The proof is presented in Sec.~\ref{subsec:proof_thm1}.
One important consequence is that this information-theoretic characterization gives us a way to select models that are ``close'' to the unconstrained latent factor model.
In fact, let us parametrize $p_W(z|x)$ with a set of parameters $W \in \mathcal{W}$ and get a family of joint distributions $\mathcal{P} = \{p_W(x,z) = p(x)p_W(z|x) : W \in \mathcal{W} \}$.
By taking $p_{W^*}(x,z) \in \text{arg}\min_{p_W(x,z) \in \mathcal{P}} TC(Z) + TC(X|Z)$ we select a joint distribution that is as close as possible to satisfy the conditional independence statements corresponding to the unconstrained latent factor model.
If for $p_{W^*}(x,z)$ we have $TC(Z) + TC(X|Z) = 0$, then by Prop.~\ref{thm1} we have a model where latent variables are independent and explain all dependencies between observed variables.
Next, we define modular latent factor models (shown in Fig.~\ref{fig:schematic-b}) and bias the learning of unconstrained latent factor models towards selecting modular structures.

\begin{definition}
A joint distribution $p(x,z)$ with $p$ observed variables $X_{1:p}$ and $m$ hidden variables $Z_{1:m}$ is called modular latent factor model if it factorizes in the following way: $ \forall x, z, ~~p(x,z) = \big( \prod_{i=1}^p{p(x_i|z_{\pi_i})}\big) \big( \prod_{j=1}^m{p(z_j)} \big) $, with $\pi_i \in \{1, 2, \ldots, m\}$.
\end{definition}
The motivation behind encouraging modular structures is two-fold.
First, modular factor models are easy to interpret by grouping the observed variables according to their latent parent.
Second, modular structures are good candidates for beating the curse of dimensionality.
Imagine increasing the number of observed variables while keeping the number of latent factors fixed. Intuitively, we bring more information about latent variables, which should help us to recover the structure better.
We get another hint on this when we apply a technique from \citet{wang2010} to lower bound the sample complexity of recovering the structure of a Gaussian modular latent factor model.
We establish that the lower bound decreases as we increase $p$ keeping $m$ fixed (refer to Sec.~\ref{sec:complexity} for more details).
For more general models such as Markov random fields, the sample complexity grows like $\log p$~\cite{wang2010}.

To give an equivalent information-theoretic characterization of modular latent factor models hereafter we focus our analysis on multivariate Gaussian distributions.

\begin{theorem}
\label{thm2}
A multivariate Gaussian distribution $p(x,z)$ is a modular latent factor model if and only if $TC(X|Z) + TC(Z)=0$ and $\forall i, TC(Z|X_i)=0$. 
\end{theorem} 
The proof is presented in Sec.~\ref{subsec:proof_thm2}.
Besides characterizing modular latent factor models, this theorem gives us an information-theoretic criterion for selecting more modular joint distributions.
The next section describes the proposed method which uses this theorem to bias the model selection procedure towards modular solutions.

\section{Linear CorEx}
\label{sec:linearcorex}
We sketch the main steps of the derivation here while providing the complete derivation in Sec.~\ref{sec:complete_derivation}.
The first step is to define the family of joint distributions we are searching over by parametrizing $p_W(z|x)$.
If $X_{1:p}$ is Gaussian, then we can ensure $X_{1:p}, Z_{1:m}$ are jointly Gaussian by parametrizing $p_W(z_j | x) = \mathcal N(w_j^T x, \eta_j^2), ~w_j \in \mathbb{R}^p, j=1..m$, or equivalently by $z = W x + \epsilon$ with $W \in \mathbb{R}^{m\times p}, \epsilon \sim \mathcal{N}(0, \text{diag}(\eta_1^2, \ldots, \eta_m^2))$.
W.l.o.g. we assume the data is standardized so that $\la X_i \ra = 0, \la X_i^2 \ra = 1$.
Motivated by Thm.~\ref{thm2}, we will start with the following optimization problem:
\begin{equation}
    \minimize_{W}  TC(X|Z) + TC(Z) + \sum_{i=1}^p Q_i,
    \label{eq:obj_1}
\end{equation}
where $Q_i$ are regularization terms for encouraging modular solutions (i.e. encouraging solutions with smaller value of $TC(Z|X_i$).
We will later specify this regularizer as a non-negative quantity that goes to zero in the case of exactly modular latent factor models.
After some calculations for Gaussian random variables and neglecting some constants, the objective simplifies as follows:
\begin{equation}
\label{eq:opt2}
\minimize_{W} \sum_{i=1}^p (\half \log \la  (X_i - \mu_{X_i|Z})^2 \ra + Q_i) + \sum_{j=1}^m \half \log \la  Z_j^2 \ra,
\end{equation}
where $\mu_{X_i|Z} = \mathbb{E}_{X_i|Z}[X_i | Z]$.
For Gaussians, calculating  $\mu_{X_i|Z}$ requires a computationally undesirable matrix inversion. 
Instead, we will select $Q_i$ to eliminate this term while also encouraging modular structure.
According to Thm.~\ref{thm2}, modular models obey $TC(Z|X_i)=0$, which implies that $p(x_i|z) = p(x_i)/p(z) \prod_j p(z_j|x_i)$.
Let $\nu_{X_i|Z}$ be the conditional mean of $X_i$ given $Z$ under such factorization. Then we have
\begin{equation*}
\nu_{X_i|Z} = \frac{1}{1+r_i} \sum_{j=1}^m \frac{Z_j B_{j,i}}{\sqrt{\la Z_j^2 \ra}}, \mbox{with }
R_{j,i} = \frac{\la X_i Z_j \ra}{\sqrt{\la X_i^2 \ra \la Z_j^2 \ra}}, 
 B_{j,i} = \frac{R_{j,i}}{1-R_{j,i}^2} 
, r_i = \sum_{j=1}^m R_{j,i} B_{j,i}.
\end{equation*}
If we let
\begin{align*}
Q_i = \frac{1}{2} \log \frac{ \la (X_i - \nu_{X_i|Z})^2 \ra}{\la (X_i - \mu_{X_i|Z})^2 \ra} = \frac{1}{2} \log\left(1+ \frac{ \la (\mu_{X_i|Z} - \nu_{X_i|Z})^2 \ra}{\la (X_i - \mu_{X_i|Z})^2 \ra}\right) \geq 0,
\end{align*}
then we can see that this regularizer is always non-negative and is zero exactly for modular latent factor models (when $\mu_{X_i|Z} = \nu_{X_i|Z}$). 
The final objective simplifies to the following:
\begin{equation}
\label{eq:opt3}
\minimize_{W} \sum_{i=1}^p \half \log \la (X_i - \nu_{X_i|Z})^2\ra + \sum_{j=1}^m \half \log \la Z_j^2 \ra.
\end{equation}
This objective depends on pairwise statistics and requires no matrix inversion.
The global minimum is achieved for modular latent factor models.
The next step is to approximate the expectations in the objective (\ref{eq:opt3}) with empirical means and optimize it with respect to the parameters $W$.
After training the method we can interpret $\hat{\pi}_i \in \text{arg}\max_{j} I(Z_j; X_i)$ as the parent of variable $X_i$.
Additionally, we can estimate the covariance matrix of $X$ the following way:
\begin{equation}
    \widehat{\Sigma}_{i, \ell \neq i} = \frac{(B^TB)_{i,\ell}}{(1 + r_i)(1 + r_\ell)} \numberthis , \ \ \
\widehat{\Sigma}_{i, i} = 1.
\label{eq:cov_formula}
\end{equation}

\begin{algorithm}[tb]
\footnotesize
\caption{\footnotesize Linear CorEx. Implementation is available at \url{https://github.com/Harhro94/T-CorEx}.}
\begin{algorithmic}
 \STATE {\bfseries Input:} {Data matrix $X \in \mathbb{R}^{n \times p}$, with $n$ iid samples of vectors in $R^p$.}
 \STATE {\bfseries Result:} {Weight matrix, $W$, optimizing (\ref{eq:opt3}).}
 \STATE Subtract mean and scale from each column of data\;
 \STATE Initialize $W_{j,i} \sim \mathcal N(0, 1 / \sqrt{p})$\;
 \FOR{$\epsilon$ in $[0.6, 0.6^, 0.6^3, 0.6^4, 0.6^5, 0.6^6, 0]$}
 \REPEAT
  \STATE $\bar{X} = \sqrt{1 - \epsilon^2} X + \epsilon E,\ \text{with } E \in \mathbb{R}^{n \times p}$ and $E_{i,j} \overset{\text{iid}}{\sim} \mathcal{N}(0,1)$ \;
  \STATE Let $\hat{J}(W)$ be the empirical version of (\ref{eq:opt3}) with $X$ replaced by $\bar{X}$
  \STATE Do one step of ADAM optimizer to update $W$ using $\nabla_W \hat{J}(W)$\;
  \UNTIL{until convergence or maximum number of iterations is reached}
  \ENDFOR
  \end{algorithmic}
  \label{alg:linearcorex}
\end{algorithm}

We implement the optimization problem (\ref{eq:opt3}) in PyTorch and optimize it using the ADAM optimizer~\cite{kingma2014adam}. In empirical evaluations, we were surprised to see that this update worked better for identifying weak correlations in noisy data than for very strong correlations with little or no noise.
We conjecture that noiseless latent factor models exhibit stronger curvature in the optimization space leading to sharp, spurious local minima.
We implemented an annealing procedure to improve results for nearly deterministic factor models.
The annealing procedure consists of rounds, where at each round we pick a noise amount, $\epsilon \in [0,1]$, and in each iteration of that round replace $X$ with its noisy version, $\bar{X}$, computed as follows: $\bar{X} = \sqrt{1-\epsilon^2} X + \epsilon E,\ \text{with } E \sim \mathcal{N}(0, I_p)$.
It can be easily seen that when $\la X_i \ra=0$, and $\la X_i^2 \ra =1$, we get that $\la \bar{X}_i \ra=0$, $\la \bar{X}_i^2 \ra=1$, and $\la \bar{X}_i \bar{X}_j \ra = (1-\epsilon^2)\la X_i X_j \ra + \epsilon^2 \delta_{i,j}$.
This way adding noise weakens the correlations between observed variables.
We train the objective (\ref{eq:opt3}) for the current round, then reduce $\epsilon$ and proceed into the next round retaining current values of parameters.
We do 7 rounds with the following schedule for $\epsilon$, $[0.6^1, 0.6^2,\ldots, 0.6^6, 0]$.
The final algorithm is shown in Alg.~\ref{alg:linearcorex}.
Our implementation is available at \url{https://github.com/Harhro94/T-CorEx}.

The only hyperparameter of the proposed method that needs significant tuning is the number of hidden variables, $m$.
While one can select it using standard validation procedures, we observed that it is also possible to select it by increasing $m$ until the gain in modeling performance, measured by log-likelihood, is insignificant.
This is due to the fact that setting $m$ to a larger value than needed has no effect on the solution of problem (\ref{eq:opt3}) as the method can learn to ignore the extra latent factors.

The stepwise computational complexity of linear CorEx is dominated by matrix multiplications of an $m \times p$ weight matrix and a $p \times n$ data matrix, giving a computational complexity of $O(m n p)$.
This is only linear in the number of observed variables assuming $m$ is constant, making it an attractive alternative to standard methods, like GLASSO, that have at least cubic complexity.
Furthermore, one can use GPUs to speed up the training up to 10 times.
The memory complexity of linear CorEx is $O((mT + n)p)$.
Fig.~\ref{fig:scaling} compares the scalability of the proposed method against other methods.

\section{Experiments}
\label{sec:experiments}
In this section we compare the proposed method against other methods on two tasks: learning the structure of a modular factor model (i.e. clustering observed variables) and estimation of covariance matrix of observed variables.
Additionally, we demonstrate that linear CorEx scales to high-dimensional datasets and finds meaningful patterns.
We present the essential details on experiments, baselines, and hyperparameters in the main text.
The complete details are presented in the appendix (see Sec.~\ref{sec:implementation_details}).

\subsection{Evidence of blessing of dimensionalty}
\label{subsec:blessing_evidence}
We start by testing whether modular latent factor models allow better structure recovery as we increase dimensionality.
We generate $n=300$ samples from a modular latent factor model with $p$ observed variables, $m=64$ latent variables each having $p/m$ children, and additive white Gaussian noise channel from parent to child with fixed signal-to-noise ratio $s=0.1$.
By setting $s=0.1$ we focus our experiment in the regime where each individual variable has low signal-to-noise ratio.
Therefore, one should expect poor recovery of the structure when $p$ is small.
In fact, the sample complexity lower bound of Thm.~\ref{thm:sample} tells us that in this setting any method needs at least 576 observed variables for recovering the structure with $\epsilon=0.01$ error probability.
As we increase $p$, we add more weakly correlated variables and the overall information that $X$ contains about $Z$ increases.
One can expect that some methods will be able to leverage this additional information. 

\begin{figure}[t!]
   \centering
   \begin{subfigure}{0.49\textwidth}
   \includegraphics[width=\columnwidth]{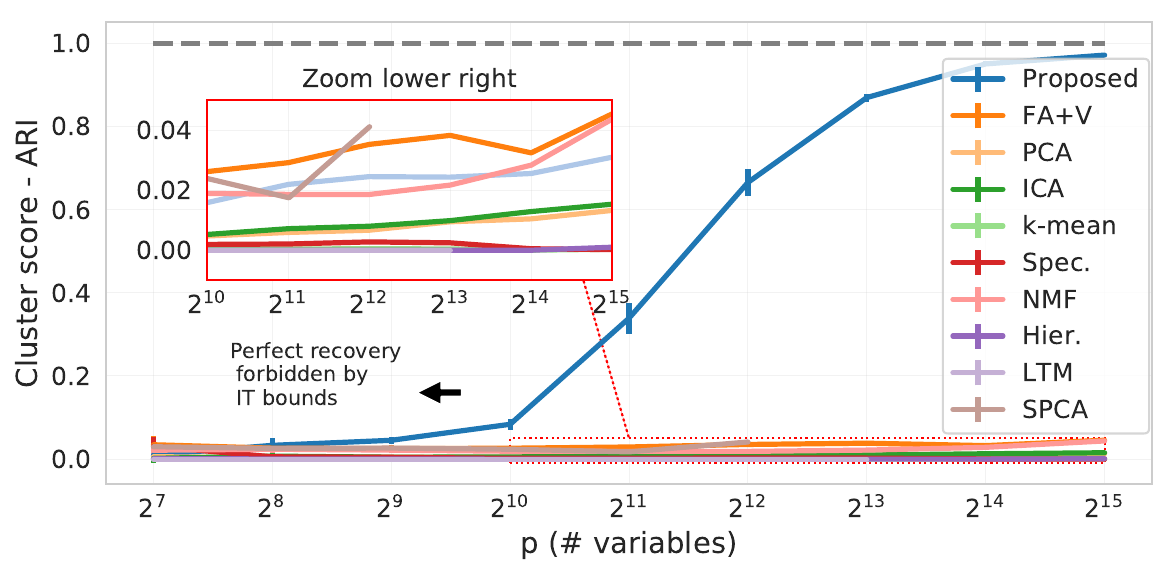}
   \end{subfigure}%
   ~
   \begin{subfigure}{0.49\textwidth}
   \includegraphics[width=\columnwidth]{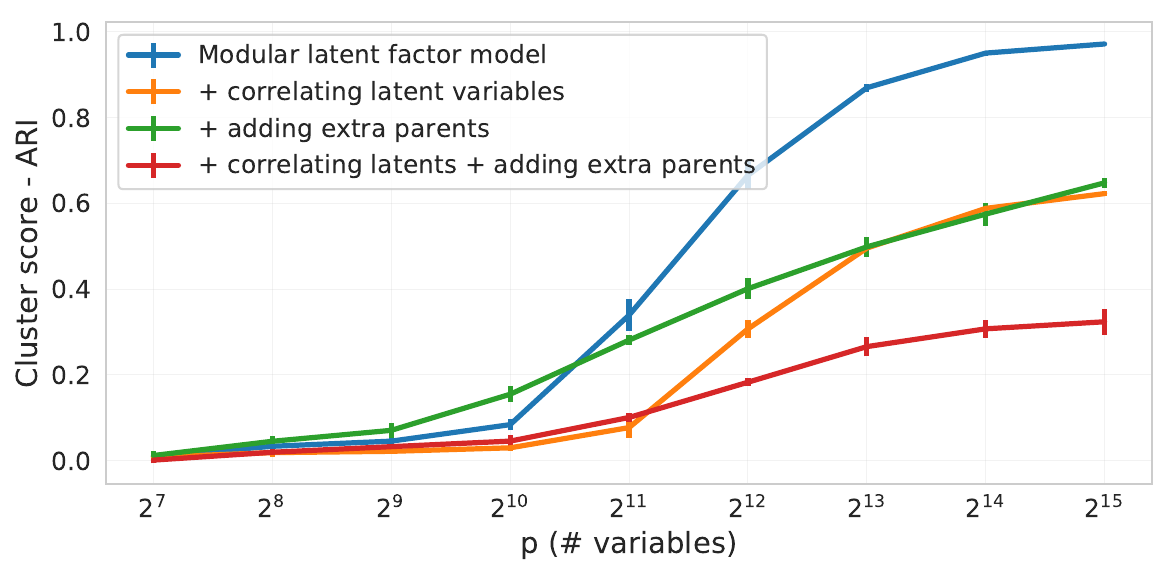}
   \end{subfigure}
   \caption{Evidence of blessing of dimensionality effect when learning modular (on the left) or approximately modular (on the right) latent factor models.
   We report adjusted Rand index (ARI) measured on $10^4$ test samples. 
   Error bars are standard deviation over 20 runs.
    }
   \label{fig:blessing}
\end{figure}

As recovering the structure corresponds to correctly clustering the observed variables, we consider various clustering approaches.
For decomposition approaches like factor analysis (FA)~\cite{factoranalysis}, non-negative matrix factorization (NMF), probabilistic principal component analysis (PCA)~\cite{ppca}, sparse PCA~\cite{zou2006sparse,mairal2009online} and independent component analysis (ICA), we cluster variables according to the latent factor whose weight has the maximum magnitude.
As factor analysis suffers from an unidentifiability problem, we do varimax rotation (FA+V) ~\cite{varimax} to find more meaningful clusters.
Other clustering methods include k-means, hierarchical agglomerative clustering using Euclidean distance and the Ward linkage rule (Hier.), and spectral clustering (Spec.)~\cite{von2007tutorial}.
Finally, we consider the latent tree modeling (LTM) method~\cite{choi_tree}.
Since information distances are estimated from data, we use the ``Relaxed RG'' method. We slightly modify the algorithm to use the same prior information as other methods in the comparison, namely, that there are exactly $m$ groups and observed nodes can be siblings, but not parent and child.
We measure the quality of clusters using the adjusted Rand index (ARI), which is adjusted for chance to give 0 for a random clustering and 1 for a perfect clustering. 
The left part of Fig.~\ref{fig:blessing} shows the clustering results for varying values of $p$.
While a few methods marginally improve as $p$ increases, only the proposed method approaches perfect reconstruction.

We find that this blessing of dimensionality effect persists even when we violate the assumptions of a modular latent factor model by correlating the latent factors or adding extra parents for observed variables.
For correlating the latent factors we convolve each $Z_i$ with two other random latent factors.
For adding extra parents, we randomly sample $p$ extra edges from a latent factor to a non-child observed variable. 
By this we create on average one extra edge per each observed variable.
In both modifications to keep the the notion of clusters well-defined, we make sure that each observed variable has higher mutual information with its main parent compared to other factors.
All details about synthetic data generation are presented in Sec.~\ref{sec:syndatasets}.
The right part of the Fig.~\ref{fig:blessing} demonstrates that the proposed method improves the results as $p$ increases even if the data is not from a modular latent factor model.
This proves that our regularization term for encouraging modular structures is indeed effective and leads to such structures (more evidence on this statement are presented in Sec.~\ref{sec:modularitychecking}).

\subsection{Covariance estimation}
We now investigate the usefulness of our proposed approach for estimating covariance matrices in the challenging undersampled regime where $n \ll p$.
For comparison, we include the following baselines: the empirical covariance matrix, Ledoit-Wolf (LW) method~\cite{ledoit2004well}, factor analysis (FA), sparse PCA, graphical lasso (GLASSO), and latent variable graphical lasso (LVGLASSO).
To measure the quality of covariance matrix estimates, we evaluate the Gaussian negative log-likelihood on a test data.
While the Gaussian likelihood is not the best evaluation metric for non-Gaussian data, we would like to note that our comparisons of baselines are still fair, as most of the baselines, such as [latent variable] GLASSO, [sparse] PCA, are derived under Gaussian assumption.
In all experiments hyper-parameters are selected from a grid of values using a 3-fold cross-validation procedure.

\textbf{Synthetic data \quad}
\begin{figure}[t!]
   \centering
   \begin{subfigure}{0.49\textwidth}
   \includegraphics[width=\textwidth]{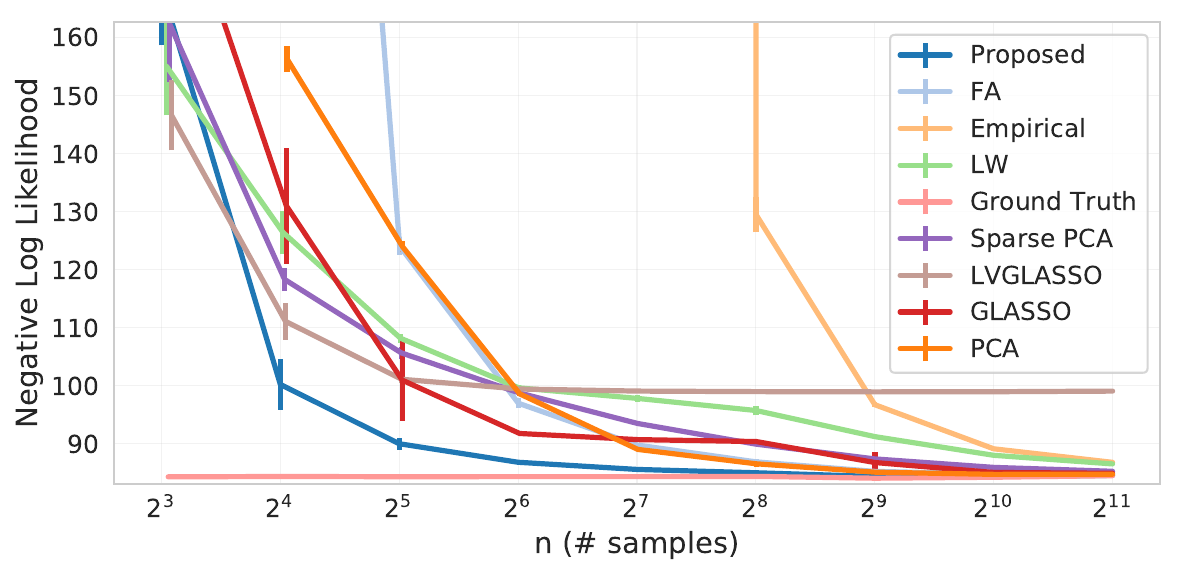}
   \end{subfigure}%
   ~
   \begin{subfigure}{0.49\textwidth}
   \includegraphics[width=\textwidth]{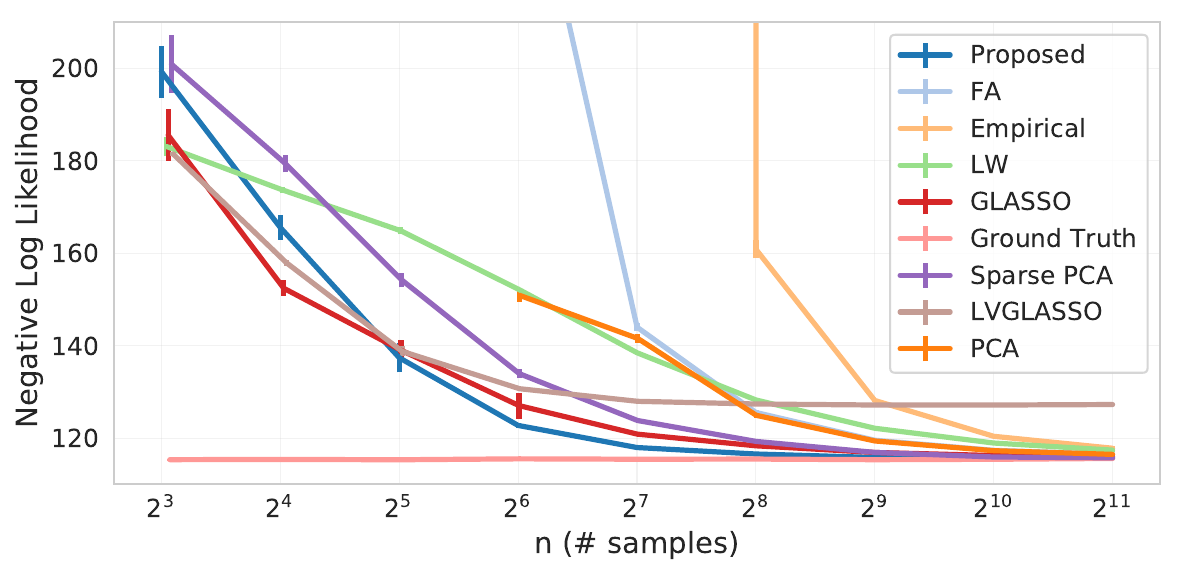}
   \end{subfigure}
   \caption{
   Comparison of covariance estimation baselines on synthetic data coming from modular latent models. On the left: $m=8$ latent factors each having $16$ children, on the right: $m=32$ latent factors each having $4$ children. The reported score is the negative log-likelihood (lower better) on a test data with 1000 samples. Error bars are standard deviation over 5 runs. We jitter $x$-coordinates to avoid overlaps.}
   \label{fig:covariance}
\end{figure}
We first evaluate covariance estimation on synthetic data sampled from a modular latent factor model.
For this type of data, the ground truth covariance matrix is block-diagonal with each block being a diagonal plus rank-one matrix.
We consider two cases: 8 large groups with 16 variables in each block and 32 small groups with 4 variables in each block.
In both cases we set the signal-to-noise ratio $\snr=5$ and vary the number of samples.
The results for both cases are shown in Fig.~\ref{fig:covariance}.
As expected, the empirical covariance estimate fails when $n \leq p$.
PCA and factor analysis are not competitive in cases when $n$ is small, while LW nicely handles those cases.
Methods with sparsity assumptions: sparse PCA, GLASSO, LVGLASSO, do well especially for the second case, where the ground truth covariance matrix is very sparse.
In most cases the proposed method performs best, only losing narrowly when $n \leq 16$ samples and the covariance matrix is very sparse.

\textbf{Stock market data \quad}
\begin{figure}[t!]
   \centering
   \begin{minipage}{0.49\textwidth}
   \includegraphics[width=\textwidth]{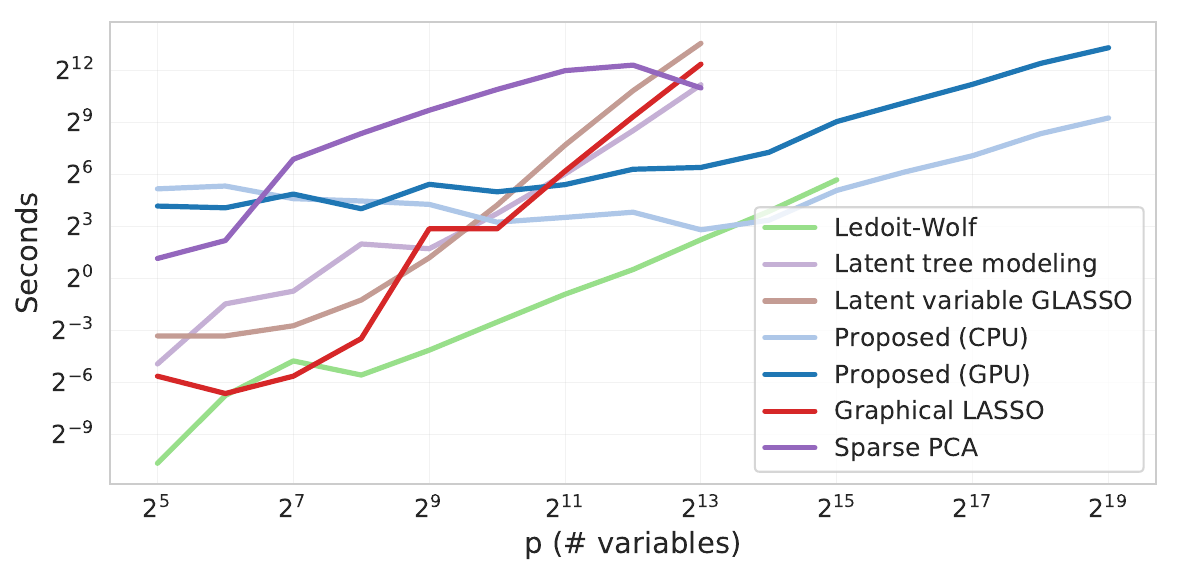}
   \caption{Runtime comparison of various methods. Points that do not appear either timed out at $10^4$ seconds or ran out of memory. The experiment was done in the setting of Sec.~\ref{subsec:blessing_evidence} on an Intel Core i5 processor with 4 cores at 4Ghz and 64Gb memory. We used Nvidia RTX 2080 GPU when running the proposed method on a GPU.} 
   \label{fig:scaling}
   \end{minipage}
   \hfill
   \begin{minipage}{0.49\textwidth}
   \includegraphics[width=\textwidth]{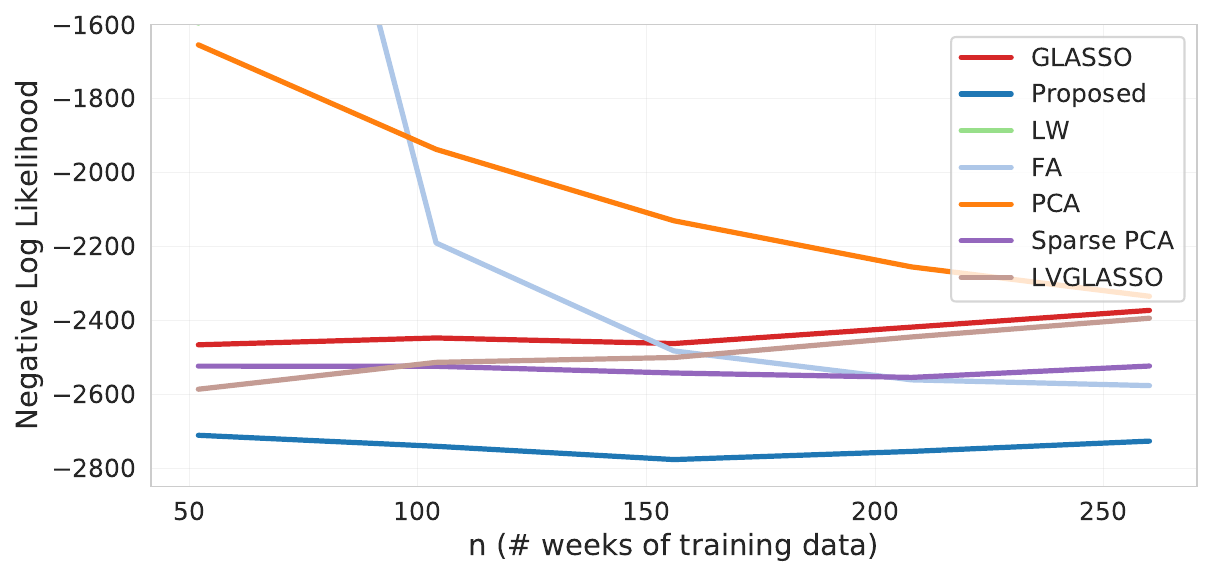} 
   \caption{Comparison of covariance estimation baselines on stock market data.
   The reported score is the negative log-likelihood (lower better) on a test data.
   Most of the Ledoit-Wolf points are above the top of the $y$ axis.}
   \label{fig:stock}
   \end{minipage}
\end{figure}
In finance, the covariance matrix plays a central role for estimating risk and this has motivated many developments in covariance estimation.
Because the stock market is highly non-stationary, it is desirable to estimate covariance using only a small number of samples consisting of the most recent data. 
We considered the weekly percentage returns for U.S. stocks from January 2000 to January 2017 freely available on \url{http://quandl.com}.
After excluding stocks that did not have returns over the entire period, we were left with 1491 companies. 
We trained on $n$ weeks of data to learn a covariance matrix using various methods then evaluated the negative log-likelihood on the subsequent 26 weeks of test data.
Each point in Fig.~\ref{fig:stock} is an average from rolling the training and testing sets over the entire time period.
For component-based methods (probabilistic PCA, sparse PCA, FA, proposed method) we used 30 components.
We omitted empirical covariance estimation since all cases have $n < p$.
We see that Ledoit-Wolf does not help much in this regime.
With enough samples, PCA and FA are able to produce competitive estimates.
Methods with sparsity assumptions, such as GLASSO, LVGLASSO, and sparse PCA, perform better.
We see that LVGLASSO consistently outperforms GLASSO, indicating that stock market data is better modeled with latent factors.
The proposed method consistently outperforms all the other methods.
Our approach leverages the high-dimensional data more efficiently than standard factor analysis. 
The stock market is not well modeled by sparsity, but attributing correlations to a small number of latent factors appears to be effective.

To examine the interpretability of learned latent factors, we used weekly returns from January 2014 to January 2017 for training. This means we used only 156 samples and 1491 variables (stocks).
For each factor, we use the mutual information between a latent factor and stock to rank the top stocks related to a factor.
We summarize the top stocks for other latent factors in Table \ref{tab:stock}.
Factor $0$ appears to be not just banking related, but more specifically bank holding companies. Factor $5$ has remarkably homogeneous correlations and consists of energy companies. Factor $9$ is specific to home construction.


\begin{table}[t!]
\caption{For the first ten latent factors, we give the top three stocks ranked by mutual information between stock and associated latent factor.}\label{tab:stock}
\centering
\scalebox{0.75}{
\begin{tabular}{@{}rll@{}} \toprule
Factor & Stock ticker & Sector/Industry \\ \midrule
0 & RF, KEY, FHN& Bank holding (NYSE, large cap)\\
1 & ETN, IEX, ITW& Industrial machinery\\
2 & GABC, LBAI, FBNC& Bank holding (NASDAQ, small cap)\\
3 & SPN, MRO, CRZO& Oil \& gas \\
4 & AKR, BXP, HIW& Real estate investment trusts \\
5 & CMS, ES, XEL& Electric utilities\\
6 & POWI, LLTC, TXN& Semiconductors\\
7 & REGN, BMRN, CELG& Biotech pharmaceuticals\\
8 & BKE, JWN, M& Retail, apparel\\
9 & DHI, LEN, MTH& Homebuilders\\
\bottomrule
\end{tabular}
}
\end{table}

\textbf{OpenML datasets \quad}
To demonstrate the generality of our approach, we show results of covariance estimation on 51 real-world datasets.
To avoid cherry-picking, we selected datasets from OpenML~\cite{openml} according to the following criteria: between $100$ and $11000$ numeric features, at least twenty samples but fewer samples than features (samples with missing data were excluded), and the data is not in a sparse format.
These datasets span many domains including gene expression, drug design, and mass spectrometry. 
For factor-based methods including our own, we chose the number of factors from the set $m \in \{5,  20, 50, 100\}$ using 3-fold cross-validation.
We use an 80-20 train-test split, learning a covariance matrix from training data and then reporting the negative log-likelihood on test data. 
We standardized the data columns to have zero mean and unit variance. Numerical problems involving infinite log-likelihoods can arise in datasets which are low rank because of duplicate columns, for example. We add Gaussian noise with variance $10^{-12}$ to avoid this.

We compared the same methods as before with three changes.
We omitted empirical covariance estimation since all cases have $n < p$.
We also omitted LVGLASSO as it was too slow on datasets having about $10^4$ variables.
The standard GLASSO algorithm was also far too slow for these datasets.
Therefore, we used a faster version called BigQUIC~\cite{bigquic}.
For GLASSO, we considered sparsity hyper-parameters $\lambda \in \{2^0, 2^1, 2^2, 2^3\}$.
We intended to use a larger range of sparsity parameters but the speed of BigQUIC is highly sensitive to this parameter.
In a test example with $10^4$ variables, the running time was 130 times longer if we use $\lambda = 0.5$ versus $\lambda = 1$.
Due to space limits we present the complete results in the appendix (Sec.~\ref{sec:openml_complete}, Table~\ref{tab:results}).
The proposed method clearly outperformed the other methods, getting the best score on 32 out of 51 datasets. Ledoit-Wolf also performed well, getting the best results on 18 out of 51 datasets. 
Even when the proposed method was not the best, it was generally quite close to the best score. 
The fact that we had to use relatively large sparsity parameters to get reasonable running time may have contributed to BigQUIC's poor performance.

\subsection{High-resolution fMRI data}
\label{sebsec:fmri}
\begin{figure}[t!]
    \centering
    \begin{subfigure}{0.48\textwidth}
        \centering
        \includegraphics[width=\textwidth]{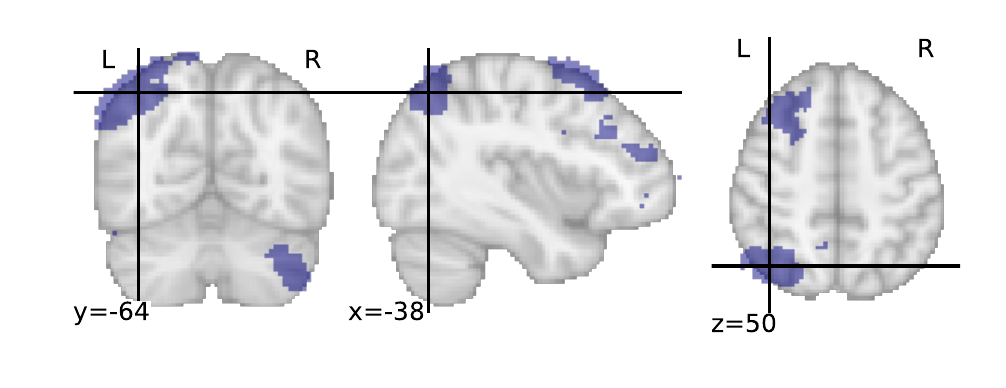}
    \end{subfigure}%
    ~
    \begin{subfigure}{0.48\textwidth}
        \centering
        \includegraphics[width=\textwidth]{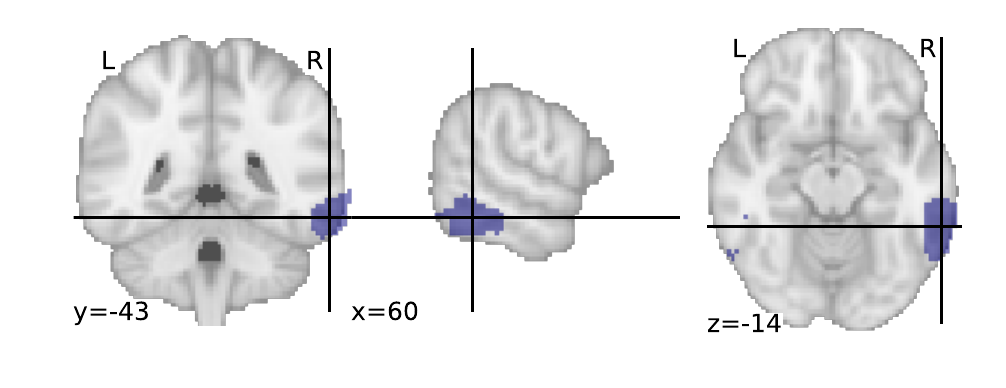}
    \end{subfigure}
    
    \begin{subfigure}{0.48\textwidth}
        \centering
        \includegraphics[width=\textwidth]{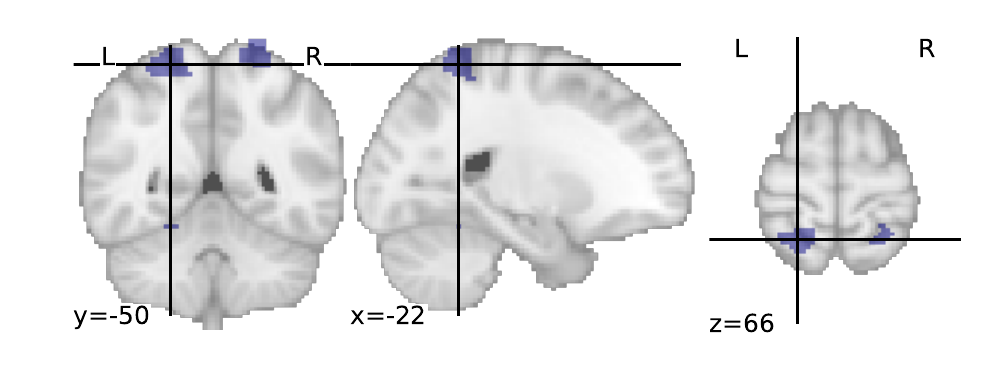}
    \end{subfigure}
    \caption{Some of the clusters linear CorEx finds. The cross-hairs correspond to the specified regions.
    }
    \label{fig:fmri-clusters}
\end{figure}
The low time and memory complexity of the proposed method allows us to apply it on extremely high-dimensional datasets, such as functional magnetic resonance images (fMRI), common in human brain mapping.
The most common measurement in fMRI is Blood Oxygen Level-Dependent (BOLD) contrast, which measures blood flow changes in biological tissues (``activation'').
In a typical fMRI session hundreds of high-resolution brain images are captured, each having 100K-600K volumetric pixels (voxels).
We demonstrate the scalability and interpretability of linear CorEx by applying it with 100 latent factors on the resting-state fMRI of the first session (session id: 014) of the publicly available MyConnectome project~\cite{poldrack2015long}.
The session has 518 images each having 148262 voxels.
We do spatial smoothing by applying a Gaussian filter with fwhm=8mm, helping our model to pick up the spatial information faster. Without spatial smoothing the training is unstable and we suspect that more samples are needed to train the model.
We assign each voxel to the latent factor that has the largest mutual information with it, forming groups by each factor.

Fig. \ref{fig:fmri-clusters} shows three clusters linear CorEx finds.
Though appearing fragmented, the cluster on the left actually captures exactly a memory and reasoning network from cognitive science literature~\cite{monti2007functional}.
This includes the activations in the Left Superior Parietal Lobule, the Left Frontal Middle and Superior Gyri, and the Right Cerebellum.
Though the authors of \cite{monti2007functional} are describing activations during a task-based experiment, the correlation of these regions during resting state is unsurprising if they indeed have underlying functional correlations.
The cluster in the middle is, with a few outlier exceptions, a contiguous block in the Right Medial Temporal cortex.
This demonstrates the extraction of lateralized regions.
The cluster on the right is a bilateral group in the Superior Parietal Lobules.
Bilateral function and processing is common for many cortical regions, and this demonstrates the extraction of one such cluster.

\section{Related work}
\label{sec:related}
Pure one factor models induce relationships among observed variables that can be used to detect latent factors~\cite{silva2,kummerfeld}. Tests using relationships among observed variables to detect latent factors have been adapted to the modeling of latent trees~\cite{choi_tree,tree2017}. 
Besides tree-like approaches or pure one factor models, another line of work imposes sparsity on the connections between latent factors and observed variables~\cite{zou2006sparse,mairal2009online}.
Another class of latent factor models can be cast as convex optimization problems~\cite{parrilo_latent,llgvm}. Unfortunately, the high computational complexity of these methods make them completely infeasible for the high-dimensional problems considered in this work. 

While sparse methods and tractable approximations have enjoyed a great deal of attention~\cite{friedman2008sparse, meinshausen2006high, CLIME, hsieh2014quic, bigquic, liu2013, lanl2017},
marginalizing over a latent factor model does not necessarily lead to a sparse model over the observed variables. Many highly correlated systems, like financial markets~\cite{fan2008high}, seem better modeled through a small number of latent factors. The benefit of adding more variables for learning latent factor models is also discussed in \cite{li2018embracing}.

Learning through optimization of information-theoretic objectives has a long history focusing on mutual information~\cite{linsker,bell95,tishby}. Minimizing $TC(Z)$ is well known as ICA~\cite{comon, ica}. The problem of minimizing $TC(X|Z)$ is less known but related to the Wyner common information~\cite{wyner_common} and has also been recently investigated as an optimization problem~\cite{gastpar}.
A similar objective was used in~\cite{corex_theory} to model \emph{discrete} variables, and a nonlinear version for continuous variables but without modularity regularization (i.e. only $TC(Z) + TC(X|Z)$) was used in \cite{corex_ae}.

\section{Conclusion}
\label{suc:conculsion}

By characterizing a class of structured latent factor models via an information-theoretic criterion, we were able to design a new approach for structure learning that outperformed standard approaches while also reducing stepwise computational complexity from cubic to linear.
Better scaling allows us to apply our approach to very high-dimensional data like full-resolution fMRI, recovering biologically plausible structure thanks to our inductive prior on modular structure.
A bias towards modular latent factors may not be appropriate in all domains and, unlike methods encoding sparsity priors (e.g., GLASSO), our approach leads to a non-convex optimization and therefore no theoretical guarantees. Nevertheless, we demonstrated applicability across a diverse set of over fifty real-world datasets, with especially promising results in domains like gene expression and finance where we outperform sparsity-based methods by large margins both in solution quality and computational cost. 

\subsubsection*{Acknowledgments}
We thank Andrey Lokhov, Marc Vuffray, and Seyoung Yun
for valuable conversations about this work and we thank
anonymous reviews whose comments have greatly improved this manuscript. H. Harutyunyan is supported by USC Annenberg Fellowship. This work is supported in part by DARPA via W911NF-16-1-0575 and W911NF-17-C-0011, and the Office of the Director of National Intelligence (ODNI), Intelligence Advanced Research Projects Activity (IARPA), via 2016-16041100004. The views and conclusions contained herein are those of the authors and should not be interpreted as necessarily representing-the official policies, either expressed or implied, of DARPA, ODNI, IARPA, or the U.S. Government. The U.S. Government is authorized to reproduce and distribute reprints for governmental purposes notwithstanding any copyright annotation therein. 

{\small \bibliography{main}}

\appendix
\clearpage
\begin{center}
\textbf{\large Supplementary material: fast structure learning with modular regularization}
\end{center}
\vskip 2em

\section{Proofs}

\subsection{Proof of proposition \ref{thm1}}
\label{subsec:proof_thm1}

\begin{proposition*}{\textbf{\ref{thm1} restated.}}
The random variables $X$ and $Z$ are described by a directed graphical model where the parents of $X$ are in $Z$ and the $Z$'s are independent if and only if $TC(X|Z) + TC(Z)=0$. 
\end{proposition*}

\begin{proof}
Because $TC$ is always non-negative, 
\begin{equation*}
TC(X|Z) + TC(Z)=0 \Leftrightarrow TC(Z)=0 \mbox{ and } TC(X|Z)=0. 
\end{equation*}
We also have the following standard statements~\cite{cover}
\begin{align*}
TC(X|Z) = 0  &\Leftrightarrow  \forall x,z, ~~p(x|z)  = \prod_{i=1}^p p(x_i|z), \\
TC(Z) = 0 &\Leftrightarrow \forall z, ~~ p(z)  = \prod_{j=1}^m p(z_j).
\end{align*}
Putting these together, we have 
\begin{equation*}
\forall x,z, ~~p(x, z)  = \prod_{i=1}^p \prod_{j=1}^m p(x_i|z) p(z_j).
\end{equation*} 
We can see that this statement is equivalent to the definition of a Bayesian network for random variables $X,Z$ with respect to the graph in Fig.~\ref{fig:schematic-a}. 
\end{proof}

\subsection{Proof of theorem \ref{thm2}}
\label{subsec:proof_thm2}
\begin{theorem*}{\textbf{\ref{thm2} restated.}}
A multivariate Gaussian distribution $p(x,z)$ is a modular latent factor model if and only if $TC(X|Z) + TC(Z)=0$ and $\forall i, TC(Z|X_i)=0$. 
\end{theorem*} 

\begin{proof}
First we show that for any modular latent factor model, even non-Gaussian, the constraints are satisfied. 
Thm.~\ref{thm1} establishes that the model implies $TC(X|Z)+TC(Z)=0$. 
We must show that the additional restriction that each $X_i$ has only one parent, $Z_{\pi_i}$, implies the condition $\forall i, TC(Z|X_i)=0$.
Looking at the rules for d-separation we see that $Z_1, \ldots, Z_m$ are independent conditioned on $X_i$. Therefore, $\forall i, TC(Z | X_i) = 0$.

Now, we show that a multivariate Gaussian distribution $p(x,z)$ with $TC(X|Z)+TC(Z)=0$ and $\forall i, TC(Z|X_i)=0$ is a modular latent factor model:
\begin{equation*}
    \forall x,z, ~~p(x,z) = \prod_{i=1}^p p(x_i|z_{\pi_i}) \prod_{j=1}^m p(z_j), 
    ~~ \text{for some } \pi_i \in \{1,2,\ldots,m\}.
\end{equation*}
By Thm.~\ref{thm1} we  have that $\forall x, z, ~~p(x,z) =  \prod_{i=1}^p p(x_i|z) \prod_{j=1}^m p(z_j)$.
To complete the proof we show that $(TC(Z) = 0 ~\&~ TC(Z|X_i)=0) \Rightarrow p(x_i |z) = p(x_i |z_{\pi_i})$ for some $\pi_i \in \{1,\ldots,m\}$. We have
\begin{align*}
\label{eq:proof2step2}
p(x_i|z) &= p(x_i)/ p(z) \prod_{j=1}^m p(z_j|x_i)  \\
&= p(x_i) \prod_{j=1}^m p(z_j|x_i) / p(z_j) \\
& =  p(x_i) \prod_{j=1}^m p(z_j, x_i) / (p(x_i) p(z_j))\numberthis.
\end{align*}
We also have that $TC(Z|X_i)=0 \Rightarrow \forall j \neq k, \text{Cov}[Z_j, Z_k | X_i] = 0$.
For Gaussians $\text{Cov}[Z_j, Z_k | X_i] = \text{Cov}[Z_j, Z_k] - \text{Cov}[Z_j, X_i] \text{Cov}[Z_k, X_i] / \text{Var}[X_i]$.
Having $\text{Var}[X_i]>0$ and $(TC(Z)=0 \Rightarrow \text{Cov}[Z_j, Z_k] = 0)$, we get $\text{Cov}[Z_j, X_i] = 0 \vee \text{Cov}[Z_k, X_i] = 0$.
Therefore, for all but at most one index, $\pi_i$, it must be the covariance of $X_i$ and $Z_j$ is zero, so that $p(z_j, x_i) = p(x_i) p(z_j)$. Putting this in Eq.~(\ref{eq:proof2step2}) we get $p(x_i|z) = p(x_i |z_{\pi_i})$. 

Note that we cannot remove the Gaussian assumption, since it is possible to have $TC(X|Z)=0, TC(Z)=0$, and $\forall i, TC(Z|X_i)=0$, but still have two non-trivial parents for one $X_i$. For example, if $Z_1, Z_2 \overset{iid}{\sim} \text{Bernoulli}(\half)$ and $X_1 = 2 Z_1 + Z_2$. It can be easily seen that the conditions are satisfied, but it is impossible to model $X_1$ with only $Z_1$ or $Z_2$ as its parent.
\end{proof}

\section{Complete derivation of linear CorEx}
\label{sec:complete_derivation}
In this section we describe the complete derivation of linear CorEx.
The first step is to define the family of joint distributions we are searching over by parametrizing $p_W(z|x)$.
If $X_{1:p}$ is Gaussian, then we can ensure $X_{1:p}, Z_{1:m}$ are jointly Gaussian by parametrizing $p_W(z_j | x) = \mathcal N(w_j^T x, \eta_j^2), ~w_j \in \mathbb{R}^p, j=1..m$, or equivalently by $z = W x + \epsilon$ with $W \in \mathbb{R}^{m\times p}, \epsilon \sim \mathcal{N}(0, \text{diag}(\eta_1^2, \ldots, \eta_m^2))$.
The noise variances $\eta_j^2$ are taken to be constants.
Please note the implicit conditional independence assumption, $TC(Z|X)=0$, we are making using this parameterization.
We do this assumption since modular latent factor models have $TC(Z|X)=0$, and it simplifies further derivations.
W.l.o.g. we assume the data is standardized so that $\la X_i \ra = 0, \la X_i^2 \ra = 1$.\footnote{Unless specified all expectations are taken with respect to the joint distribution $p_W(x,z).$}
If it is not standardized we can standardize it using the empirical means and standard deviations.
Motivated by Thm.~\ref{thm2}, we will start with the following optimization problem:
\begin{equation}
    \minimize_{W}  TC(X|Z) + TC(Z) + \sum_{i=1}^p Q_i,
    \label{eq:app_obj_1}
\end{equation}
where $Q_i$ are regularization terms for encouraging modular solutions (i.e. encouraging solutions with smaller value of $TC(Z|X_i$).\footnote{One can set $Q_i \propto TC(Z|X_i)$. However, we choose not to do this since we do not have a derivation that leads the resulting objective into an equivalent, but efficiently computable objective.}
We will later specify this regularizer as a non-negative quantity that goes to zero in the case of exactly modular latent factor models.
The $TC(X|Z) + TC(Z)$ part of the Eq.~(\ref{eq:app_obj_1}) can be rewritten as follows:
\begin{align*}
TC(X|Z) + TC(Z) & = \sum_{i=1}^p H(X_i|Z) - H(X|Z) + \sum_{j=1}^m H(Z_j) - H(Z) \\ 
& = \sum_{i=1}^p H(X_i|Z) + \sum_{j=1}^m H(Z_j) - (H(X|Z) + H(Z)) \\
& = \sum_{i=1}^p H(X_i|Z) + \sum_{j=1}^m H(Z_j) - (H(Z|X) + H(X)) \\
& = \sum_{i=1}^p H(X_i|Z) + \sum_{j=1}^m (H(Z_j) - H(Z_j|X)) + H(X) \\
& \propto \sum_{i=1}^p H(X_i|Z) + \sum_{j=1}^m I(Z_j;X) \numberthis.
\label{eq:app_eq_2}
\end{align*}
The first two lines invoke definitions and re-arrange. 
The third line uses Bayes' rule to rewrite the entropies.
The fourth line invokes conditional independence of $Z$'s conditioned on X.
Next, we write out the explicit form of expressions in Eq.~(\ref{eq:app_eq_2}) for Gaussians and ignore constants:
\begin{align*}
\lefteqn{\sum_{i=1}^p H(X_i|Z) + \sum_{j=1}^m I(Z_j;X)}\\
&= \sum_{i=1}^p \frac{1}{2} \mathbb{E}_Z \log \left(2 \pi e  \text{Var}[X_i | Z]\right) + \sum_{j=1}^m \left( H(Z_j) - H(Z_j | X) \right)\\
&= \sum_{i=1}^p \frac{1}{2} \log \mathbb{E}_Z \left[2 \pi e  \text{Var}[X_i | Z]\right] + \sum_{j=1}^m \left( H(Z_j) - H(Z_j | X) \right)\\
&\propto \frac{1}{2} \sum_{i=1}^p \log \la (X_i - \mathbb{E}_{X_i|Z}[X_i | Z])^2 \ra + \frac{1}{2}\sum_{j=1}^m \left( \log \text{Var}[Z_j] - \mathbb{E}_X \log \text{Var}[Z_j | X] \right)\\
&\propto \frac{1}{2} \sum_{i=1}^p \log \la (X_i - \mathbb{E}_{X_i|Z}[X_i | Z])^2 \ra + \frac{1}{2}\sum_{j=1}^m \left( \log \la Z_j^2 \ra - \log(\eta_j^2) \right)\\
&\propto \frac{1}{2} \sum_{i=1}^p \log \la (X_i - \mathbb{E}_{X_i|Z}[X_i | Z])^2 \ra + \frac{1}{2}\sum_{j=1}^m \log \la Z_j^2 \ra.\numberthis
\label{eq:app_eq_3}
\end{align*}
We used the fact that the differential entropy of a Gaussian variable with variance $\sigma^2$ is equal to $\half \log(2\pi e \sigma^2)$.
Also, we used the fact that if $A, B$ are jointly Gaussian random variables, then $H(A|B) \propto \mathbb{E}_B \log \text{Var}[A | B] = \log \mathbb{E}_B \text{Var}[A | B]$.
The logarithm and expectation can be swapped because for Gaussians $\text{Var}[A | B]$ is constant for any value of $B$.
In the fifth line we replace $\text{Var}[Z_j]$ with $\la Z_j^2 \ra$, because having $\la X \ra = 0$ and $z_j = w_j^T x + \epsilon_j$ implies $\la Z_j \ra = 0$.
Considering Eq.~(\ref{eq:app_eq_3}), the problem (\ref{eq:app_obj_1}) becomes:
\begin{equation}
\label{eq:app_obj_2}
\minimize_{W} \sum_{i=1}^p (\half \log \la  (X_i - \mu_{X_i|Z})^2 \ra + Q_i) + \sum_{j=1}^m \half \log \la  Z_j^2 \ra,
\end{equation}
where $\mu_{X_i|Z} = \mathbb{E}_{X_i|Z}[X_i | Z]$.
For Gaussians, calculating  $\mu_{X_i|Z}$ requires a computationally undesirable matrix inversion. 
Instead, we will select $Q_i$ to eliminate this term while also encouraging modular structure.
According to Thm.~\ref{thm2}, modular models obey $TC(Z|X_i)=0$, which implies that $p(x_i|z) = p(x_i)/p(z) \prod_j p(z_j|x_i)$.
Let $\nu_{X_i|Z}$ be the conditional mean of $X_i$ given $Z$ under such factorization.
Then it will have the following form (see Sec.~\ref{subsec:nu} for the derivation):
\begin{align*}
\nu_{X_i|Z} = \frac{1}{1+r_i} \sum_{j=1}^m \frac{Z_j B_{j,i}}{\sqrt{\la Z_j^2 \ra}},&\\
\mbox{with } R_{j,i} = \frac{\la X_i Z_j \ra}{\sqrt{\la X_i^2 \ra \la Z_j^2 \ra}}&, B_{j,i} = \frac{R_{j,i}}{1-R_{j,i}^2}, r_i = \sum_{j=1}^m R_{j,i} B_{j,i}.
\end{align*}
We see that computing $\nu_{X_i|Z}$ is easier since it requires no matrix inversion and depends only on pairwise statistics between observed and latent variables. If we let $Q_i = \half \log \la (X_i - \nu_{X_i|Z})^2 \ra - \half \log \la (X_i - \mu_{X_i|Z})^2 \ra$, we will replace $\mu_{X_i|Z}$ with $\nu_{X_i|Z}$ in problem (\ref{eq:app_obj_2}).
To see why this also encourages modular structures we note that
\begin{align*}
Q_i &= \frac{1}{2} \log \la (X_i - \nu_{X_i|Z})^2 \ra - \frac{1}{2} \log \la (X_i - \mu_{X_i|Z})^2 \ra\\
&= \frac{1}{2} \log \frac{ \la (X_i - \nu_{X_i|Z})^2 \ra}{\la (X_i - \mu_{X_i|Z})^2 \ra}\\
&= \frac{1}{2} \log\left(\frac{ \la (X_i - \nu_{X_i|Z} + \mu_{X_i|Z} - \mu_{X_i|Z})^2 \ra}{\la (X_i - \mu_{X_i|Z})^2 \ra}\right)\\
&= \frac{1}{2} \log\left(\frac{ \la (X_i - \mu_{X_i|Z})^2 \ra + \la (\mu_{X_i|Z} - \nu_{X_i|Z})^2 \ra + 2 \la (X_i - \mu_{X_i|Z})(\mu_{X_i|Z} - \nu_{X_i|Z}) \ra}{\la (X_i - \mu_{X_i|Z})^2 \ra}\right)\\
&= \frac{1}{2} \log\left(1 + \frac{ \la (\mu_{X_i|Z} - \nu_{X_i|Z})^2 \ra + 2 \mathbb{E}_Z \mathbb{E}_{X_i|Z} \left[(X_i - \mu_{X_i|Z})(\mu_{X_i|Z} - \nu_{X_i|Z}) \right]}{\la (X_i - \mu_{X_i|Z})^2 \ra}\right)\\
&= \frac{1}{2} \log\left(1+ \frac{ \la (\mu_{X_i|Z} - \nu_{X_i|Z})^2 \ra}{\la (X_i - \mu_{X_i|Z})^2 \ra}\right) \geq 0.
\end{align*}
We see that this regularizer is always non-negative and is zero exactly for modular latent factor models (when $\mu_{X_i|Z} = \nu_{X_i|Z}$). 
Summing up, the final objective simplifies to the following:
\begin{equation}
\label{eq:app_opt3}
\minimize_{W} \sum_{i=1}^p \half \log \la (X_i - \nu_{X_i|Z})^2\ra + \sum_{j=1}^m \half \log \la Z_j^2 \ra.
\end{equation}
This objective depends on pairwise statistics and requires no matrix inversion.
The global minimum is achieved for modular latent factor models.
The next step is to approximate the expectations in the objective (\ref{eq:opt3}) with empirical means and optimize it with respect to the parameters $W$.

After training the method we can interpret $\hat{\pi}_i \in \text{arg}\max_{j} I(Z_j; X_i) = \text{arg}\max_j -\half \log (1-R^2_{j,i}) = \text{arg}\max_j \lvert R_{j,i} \rvert$ as the parent of variable $X_i$.
Additionally, we can estimate the covariance matrix of the observed variables.
The method we use for estimating the covariance is as follows.
First, we have assumed that the data is standardized, so we just need to calculate the off-diagonal terms.
If $TC(X|Z)=0$, this implies the conditional covariance of $X$ given $Z$ is diagonal.
Additionally, using the law of total covariance we have:
\begin{equation*}
\text{Cov}\left[X_i, X_{\ell \neq i}\right] = \mathbb{E}\left[\text{Cov}[X_i, X_{\ell} |Z]\right] + 
\text{Cov}\left[\mu_{X_i|Z}, \mu_{X_\ell |Z}\right].
\end{equation*}
By combining the last two statements we get:
\begin{equation*}
\mathbb{E}\left[\text{Cov}[X_i, X_{\ell \neq i} |Z]\right] = \la X_i X_\ell \ra - \la \mu_{X_i|Z} \mu_{X_\ell |Z} \ra = 0.
\end{equation*}
If we assume the constraints $TC(Z)=0 ~\&~ \forall i, TC(Z|X_i)=0$ are satisfied, we saw that this implies  $\mu_{X_i|Z} = \nu_{X_i|Z}$.
Also, as $TC(Z)=0 \Rightarrow \la Z_j Z_k \ra = \delta_{j,k} \la Z_j^2 \ra$, the off-diagonal elements of $\la X_i X_\ell \ra$ satisfy:
\begin{equation*}
\la X_i X_{\ell \neq i} \ra = \la \nu_{X_i|Z} \nu_{X_\ell |Z} \ra = \frac{(B^\top B)_{i,\ell}}{(1+r_i) (1+r_\ell)}.
\end{equation*}
In conclusion we get the following covariance matrix estimates:
\begin{equation*}
    \widehat{\Sigma}_{i, \ell \neq i} = \frac{(B^TB)_{i,\ell}}{(1 + r_i)(1 + r_\ell)} \numberthis , \ \ \
\label{eq:app_cov}
\widehat{\Sigma}_{i, i} = 1.
\end{equation*}
Note that the covariance matrix estimate corresponds to the covariance matrix of the learned model if $TC(X|Z)=0, TC(Z)=0$, and $\forall i, TC(Z|X_i)=0$, i.e. the learned model is modular.
Otherwise it is an approximation of to the covariance matrix of the learned model.
From Eq.~\ref{eq:app_cov} we see that the estimates are low-rank plus diagonal matrices.
In case when the learned model is modular, it is also block-diagonal with each block being a diagonal plus rank-one matrix.
Therefore, encouraging modular structures pushes the low-rank covariance estimate to be also block-diagonal with each block being a diagonal plus rank-one matrix.

\subsection{Derivation of the conditional mean under modularity constraints}
\label{subsec:nu}

Under the conditions that $X, Z$ are jointly Gaussian and $\forall i, TC(X|Z_i)=0$, we would like to derive the mean of $X_i$ conditioned on $Z$, $\nu_{X_i|Z}$. 
We have that $TC(Z|X_i)=0 \Rightarrow p(x_i|z) = p(x_i)/p(z) \prod_j p(z_j|x_i)$. 
We will look at the distribution $q(x_i|z) = p(x_i)/p(z) \prod_j p(z_j|x_i)$ and calculate the conditional mean of this distribution. 

Let $R_{j,i}$ be the Pearson correlation coefficient between $Z_j$ and $X_i$ whose means and standard deviations are respectively indicated with $\nu_j, \rho_j$ and $\mu_i, \sigma_i$ (all with respect to the distribution $p$). 
The marginal distribution for the Gaussian distribution relating $Z_j$ and $X_i$ is well known:
\begin{equation*}
p(z_j|x_i) = \mathcal N(\nu_j + R_{j,i}\rho_j /\sigma_i (x_i - \mu_i), (1-R_{j,i}^2) \rho_j^2).
\end{equation*}

Now we look only at the exponents of $q(x_i|z)$, ignoring the normalization, to get the following:
\begin{align*}
-\log q(x_i|z) \propto (x_i - \mu_i)^2/\sigma_i^2 + \sum_{j=1}^m (z_j - \nu_j - R_{j,i}\rho_j /\sigma_i (x_i - \mu_i))^2 / ((1-R_{j,i}^2) \rho_j^2).
\end{align*}
Collecting only the terms involving $x_i$ we get the following:
\begin{align*}
-&\log q(x_i|z) \propto A x_i^2 + B x_i + C, \\
& \text{ with } A = 1/\sigma_i^2 + \sum_{j=1}^m \frac{R_{j,i}^2 \rho_j^2 /\sigma_i^2}{(1-R_{j,i}^2)\rho_j^2}, ~B = -2 \mu_i/\sigma_i^2 - \sum_{j=1}^m \frac{2(z_j - \nu_j + \mu_i R_{j,i} \rho_j / \sigma_i) R_{j,i} \rho_j / \sigma_i}{(1-R_{j,i}^2)\rho_j^2}.
\end{align*}
From completing the square, we see that the conditional mean of $X_i |Z$ has the form $\nu_{X_i|Z} = -B/(2 A)$.

Finally, we simplify the formulae because $\mu_i = \la X_i \ra = \nu_j = \la Z_j \ra = 0$ and $\sigma_i^2 = \la X_i^2 \ra = 1$. This implies that $R_{j,i} = \la X_i Z_j \ra /\sqrt{\la X_i^2 \ra \la Z_j^2 \ra}$, leaving us with the following form:
\begin{align*}
\nu_{X_i|Z} = \frac{1}{1+r_i} \sum_{j=1}^m B_{j,i} \frac{Z_j}{\sqrt{\la Z_j^2 \ra}},~~ \mbox{with } B_{j,i} = \frac{R_{j,i}}{1-R_{j,i}^2}, r_i = \sum_{j=1}^m R_{j,i} B_{j,i}.
\end{align*}

\section{Sample complexity lower bound}
\label{sec:complexity}
In this section we derive a lower bound on sample complexity for learning the structure of modular latent factor model. We follow the construction of information-theoretic sample complexity bounds in~\cite{wang2010}.
\begin{theorem}
\label{thm:sample}
For a multivariate Gaussian modular latent factor model with $p$ observed variables $X_{1:p}$, $m$ latent variables $Z_{1:m}$ with $p/m$ children each and additive white Gaussian noise channel from parent to child with signal-to-noise ratio $\snr$, the number of samples, $n$, required to recover the structure of the graphical model with error probability $\epsilon$ is lower bounded as 
\begin{equation}
\label{eq:lowerbound}
n \geq \frac{2\left((1-\epsilon) \log \left(\binom{p}{p/m,\ldots,p/m} \frac{1}{m!}\right) - 1\right)}
{(p-1) \log(1+ \snr \frac{1-1/m}{1-1/p}) - (m-1) \log (1 + \snr \frac{p}{m})}.
\end{equation}
\end{theorem}

\begin{proof}
Consider the class of modular latent factor models with $p$ observed variables and $m$ latent factors each having exactly $p/m$ children.
To distinguish the structure among this class of models corresponds to partitioning the observed variables into $m$ equally sized groups.
The number of such groupings is,
\begin{equation*}
M=\binom{p}{p/m,\ldots,p/m} \frac{1}{m!},
\end{equation*}
the multinomial coefficient for dividing $p$ items into $m$ equally sized boxes, divided by the number of indistinguishable permutations among boxes, $m!$.
We take $\theta \in \{1,\ldots,M\}$ to be an index specifying a model in this ensemble. 
Now learning the structure corresponds to finding $\theta$ from data.

W.l.o.g. assume $\forall j, \text{Var}[Z_j]=b$.
Then $X_i = Z_{\pi_{\theta}(i)} + \eta_i$, where $\pi_{\theta}(i)$ is the index of the parent of $X_i$ in model $\theta$ and $\eta_i$ is independent Gaussian noise with variance $a$. Since we have fixed the signal-to-noise ratio, we have that $a=b/s$.
W.l.o.g. we can assume that $\forall i, \mathbb{E}[X_i]=0$.
Then the covariance matrix of observed variables, $\cov_{\theta, i, j} = \mathbb{E}[X_i X_j] = b \delta_{\pi_{\theta}(i), \pi_{\theta}(j)} + a \delta_{i,j}$, where $\delta$ is the Kronecker delta.

Fano's inequality tells us that the probability of an error, $\epsilon$, in picking the correct index $\theta$ given $n$ samples of data, $X_{1:p}^{1:n}$, is bounded as follows:
\begin{equation*}
\epsilon \geq 1 - \frac{I(\theta;X_{1:p}^{1:n}) + 1}{\log M}.
\end{equation*}
Following \cite{wang2010}, we use an upper bound for the mutual information, $I(\theta;X_{1:p}^{1:n}) \leq nF/2$, where 
\begin{equation*}
F = \log \det \bar \cov - 1/M \sum_{\theta=1}^M \log \det \cov_\theta,
\end{equation*}
and ${\bar \cov = 1/M \sum_{\theta=1}^M \cov_\theta}$. 
Re-arranging Fano's inequality gives the following sample complexity bound: 
\begin{equation*}
n \geq 2 \frac{(1 - \epsilon) \log M -1}{F}.
\end{equation*}
All that remains is to find an expression for $F$. To build intuition, we explicitly write out the case for $p=4, m=2$, and for some $\theta$. 
\begin{equation*}
\cov_\theta = \left[
\begin{array}{cccc}
 b + a  & b & 0 & 0  \\ 
 b  & b + a & 0 & 0  \\ 
 0  & 0 & b + a & b  \\ 
 0 & 0 & b & b + a  \\ 
\end{array}\right]
\end{equation*}
Clearly this is a block diagonal matrix where each block is a diagonal plus rank-one (DPR1) matrix.
After we average over all $\theta$ to get $\bar \cov$, every off-diagonal entry will be the same, equal to the probability of $j \neq i$ being in the same group as $i$, or $(p/m -1) / (p-1)$.
Therefore $\bar \cov$ is also a DPR1 matrix.
Using standard identities for block diagonal and DPR1 matrices, we calculate the determinants:
\begin{align*}
\det \cov_\theta &= a^p \left(1 + \frac{b}{a} \frac{p}{m}\right)^m,\\ 
\det \bar \cov &= a^p \left(1 + \frac{b}{a} \frac{p}{m}\right) \left(1 + \frac{b}{a}\frac{p}{m} \left(\frac{m-1}{p-1} \right) \right)^{p-1}.\\
\end{align*}
Finally, we can combine all of these expressions to get a lower bound for sample complexity that depends only on $p, m$, and the signal-to-noise ratio, $\snr = b/a$. 
\end{proof}

\begin{figure}[t!]
   \centering
   \begin{subfigure}{0.49\textwidth}
   \includegraphics[width=\textwidth]{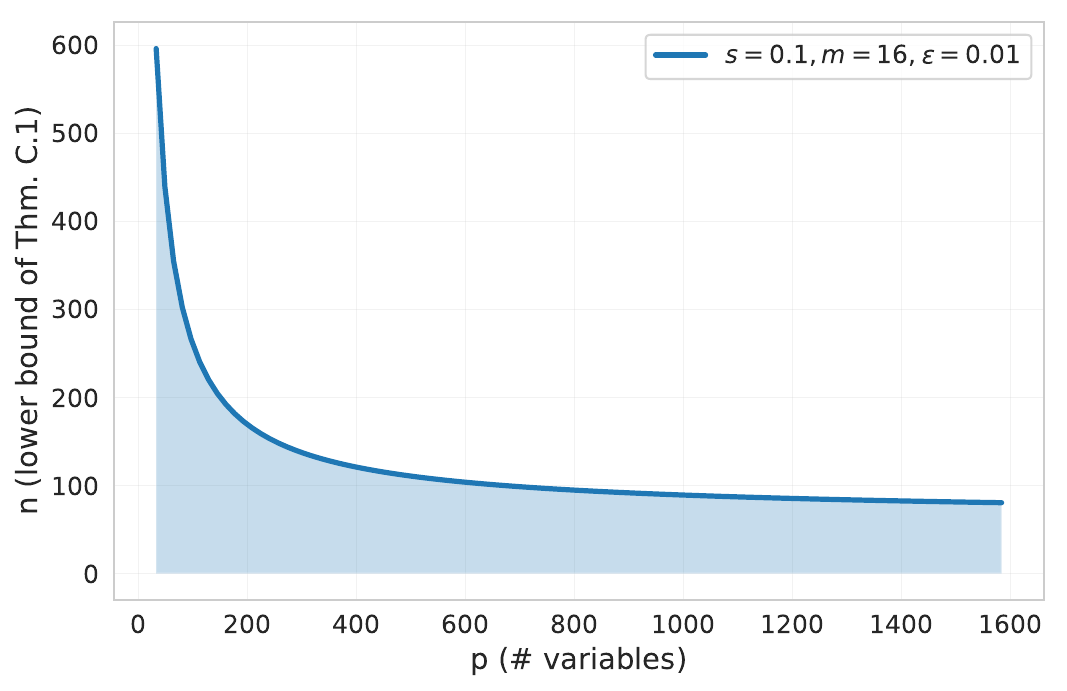}
   \end{subfigure}%
   ~
   \begin{subfigure}{0.49\textwidth}
   \includegraphics[width=\textwidth]{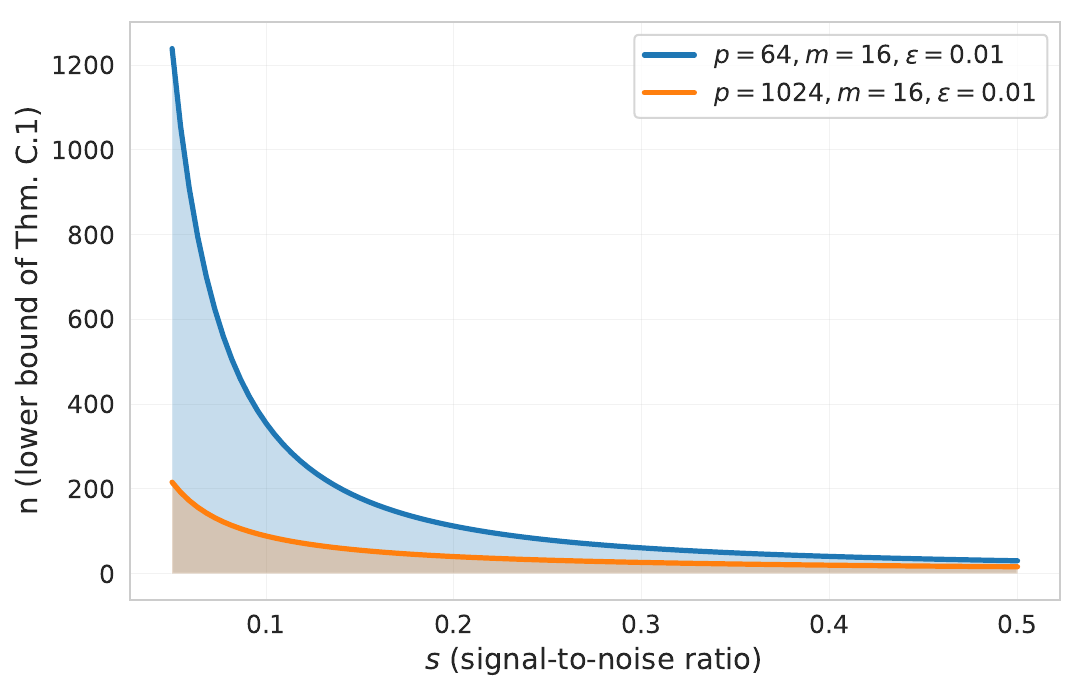}
   \end{subfigure}
   \caption{Theorem~\ref{thm:sample} prevents perfect structure recovery in the shaded region. On the left: for fixed signal-to-noise ratio and number of latent factors, the lower bound of Thm.~\ref{thm:sample} decreases as the number of observed variables increases. On the right: the same effect is visible for other values of signal-to-noise ratio.}
   \label{fig:lowerbound}
\end{figure}

The bound of Thm.~\ref{thm:sample} is not very intuitive because it involves logarithm of a multinomial coefficient.
We provide a simpler asymptotic expression for the bound.
Using Stirling's approximation we have that $\log \binom{p}{p/m,\ldots,p/m} \frac{1}{m!} \approx p \log m + 1/2 \log (p/m) - m/2 \log (m~p~ 2 \pi / e^2)$ for large values of $p$.
In the limit of large $p$, this approximation gives us the following lower bound:
\begin{equation*}
n \geq \frac{2 (1-\epsilon) \log m}{\log \left(1+\snr (1-1/m) \right)}.    
\end{equation*}
Wee see that in the limit of large $p$ the bound becomes constant rather than becoming infinite. 
Moreover, when we plot the lower bound of Eq.~(\ref{eq:lowerbound}) in Fig.~\ref{fig:lowerbound}, we see that for fixed number of latent factors the bound goes down as we increase $p$.
These two facts together hint (but do not prove) that modular latent factor models may allow blessing of dimensionality.
An evidence of blessing of dimensionality is demonstrated in Sec.~\ref{sec:experiments}.
Intuitively, recovery gets easier because more variables provide more signal to reconstruct the fixed number of latent factors.
While it is tempting to retrospectively see this as obvious, the same argument could be (mistakenly) applied to other families of latent factor models, such as the unconstrained latent factor models shown in Fig.~\ref{fig:schematic-a}, for which the sample complexity grows as we increase $p$~\cite{chandrasekaran2010latent, choi_tree}.

\section{Implementation details}
\label{sec:implementation_details}
In this section we present details on baselines, experiments, and hyperparameters.

\paragraph{Baselines} For factor analysis, PCA, sparse PCA, independent component analysis, k-means clustering, spectral clustering, negative matrix factorization, hierarchical agglomerative clustering using Euclidean distance, hierarchical agglomerative clustering the Ward linkage rule, Ledoit-Wolf and graphical LASSO we used the scikit-learn implementations~\cite{scikit-learn}.
We implemented latent tree modeling with the ``Relaxed RG'' method.
We slightly modified latent tree modeling to use the same prior information as other methods in the comparison, namely, that there are exactly m groups and observed nodes can be siblings, but not parent and child.
For latent variable graphical LASSO, we used the implementation available in the REGAIN repository.\footnote{\url{https://github.com/fdtomasi/regain}}

\paragraph{Experimental setups}
In the blessing of dimensionality experiments all methods were given the correct number of clusters.
The scores were computed using 10000 test examples.
When possible we reported the means and standard deviations over 20 runs.
In the covariance estimation experiments with synthetic data, models requiring a number of latent factors or a number of components were given the correct number.
The scores were computed using 1000 test examples.
We reported the means and standard deviations over 5 runs.
In the stock market experiments models were trained on $n$ weeks and their estimates were evaluated using the negative log-likelihood on the subsequent 26 weeks.
We presented the average score from rolling the training and testing sets over the entire time period.
Standard deviations are not presented because scores corresponding to different time periods are very different, resulting in large standard deviations.
This is due to the stock market exhibiting different behaviour in different time periods.
In experiments with OpenML datasets we used a random 80-20 train-test split.
We reported the negative log-likelihood on test sets.
As large amount of computation is needed to generate results on OpenML datasets, we did only a single run for each dataset.

\paragraph{Hyperparameters}
In all cases the proposed method was trained using Adam optimizer with $0.01$ learning rate, $\beta_1=0.9$, and $\beta_2=0.999$.
In all covariance estimation problems the hyperparameters were selected from a grid of values using a 3-fold cross-validation procedure.
The sparsity parameter of sparse PCA was selected from $[0.1, 0.3, 1.0, 3.0, 10.0]$.
The sparsity parameters of GLASSO and latent variable GLASSO were selected from $[0.01, 0.1, 0.3, 1.0, 3.0, 10.0]$.
For latent variable GLASSO, the additional regularization parameter (``tau'', controlling the nuclear norm of the low-rank part of the inverse covariance matrix) was selected from $[0.01, 0.1, 1.0, 10.0, 100.0]$.
In the experiments with OpenML datasets the sparsity hyperparameter of BigQUIC was selected form $[2^0, 2^1, 2^2, 2^3]$.
In the timing experiments the sparsity parameters of sparse PCA and GLASSO were set to $1.0$.
LVGLASSO was trained with the sparsity parameter set to $0.1$ and with ``tau'' set to $30.0$.

\section{Details on generating synthetic data}
\label{sec:syndatasets}
\begin{figure}[t!]
    \centering
    \begin{subfigure}{0.49\textwidth}
    \includegraphics[width=\textwidth]{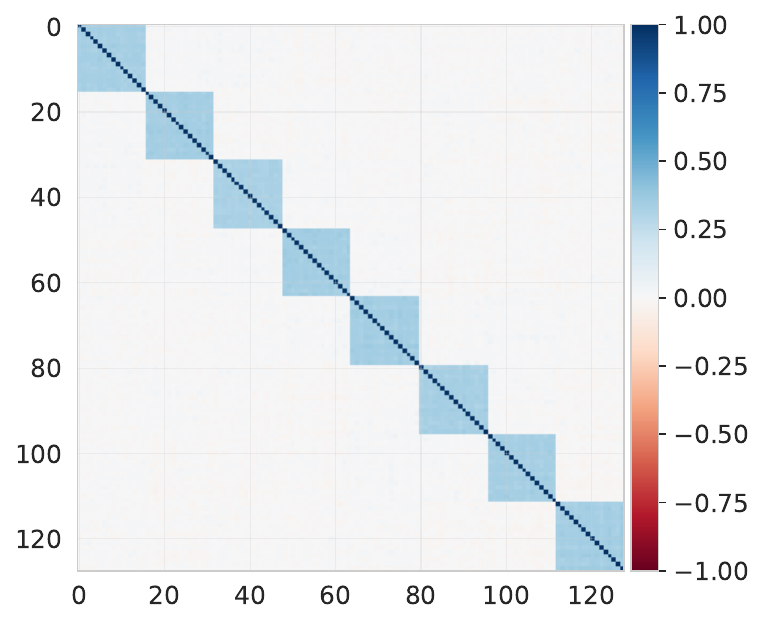}
    \caption{Modular latent factor model}
    \label{fig:cov_pure_modular}
    \end{subfigure}~
    \begin{subfigure}{0.49\textwidth}
    \includegraphics[width=\textwidth]{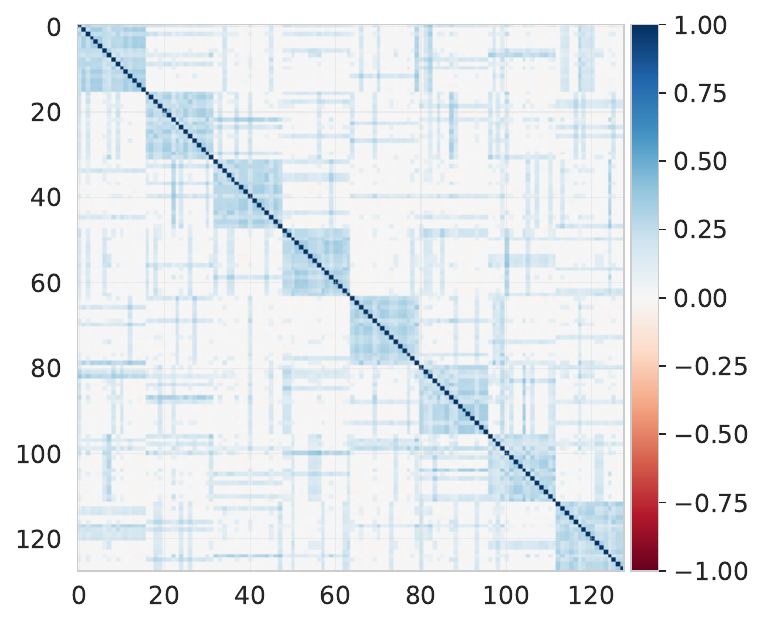}
    \caption{Modular + extra parents}
    \label{fig:cov_extraparents}
    \end{subfigure}
    ~
    \begin{subfigure}{0.49\textwidth}
    \includegraphics[width=\textwidth]{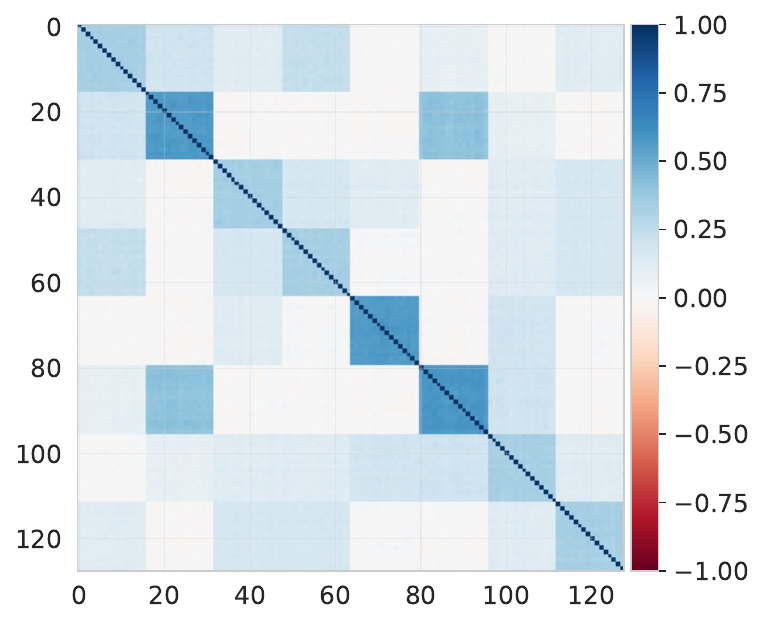}
    \caption{Modular + correlated $Z$s}
    \label{fig:cov_correlatedzs}
    \end{subfigure}~
    \begin{subfigure}{0.49\textwidth}
    \includegraphics[width=\textwidth]{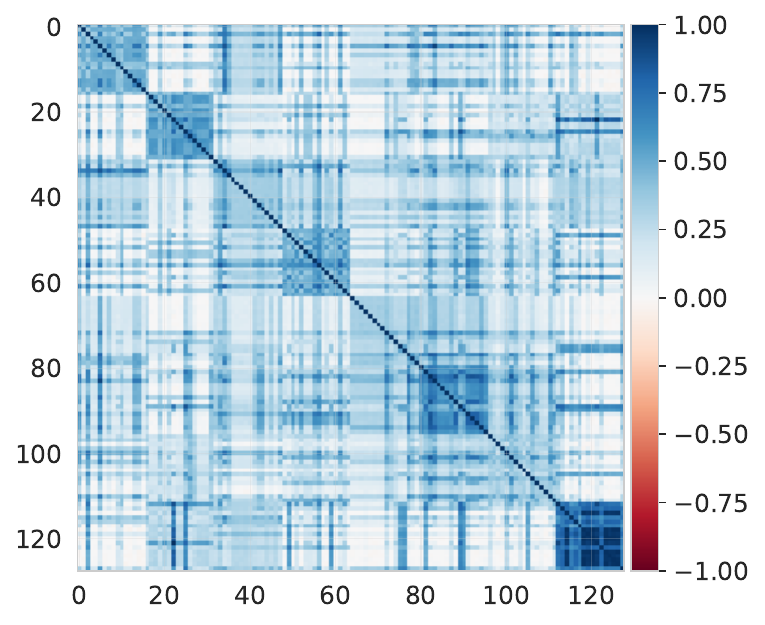}
    \caption{Modular + extra parents and correlated $Z$s}
    \label{fig:cov_extraparents_correlatedzs}
    \end{subfigure}
    \caption{Empirical covariance matrices (estimated using $n=10^4$ samples) corresponding to modular (a) and approximately modular (b, c, d) latent factor models. In all examples $m=8$, $p=128$, $\snr=0.5$.} 
    \label{fig:example_cov_matrices}
\end{figure}
In all experiments involving a synthetic modular latent factor model we generate the data the following way.
We first take $m$ independent standard Gaussian random variables, $Z_1, Z_2, \ldots, Z_m \overset{iid}{\sim} \mathcal{N}(0,1)$.
For simplicity we assume that $m$ divides $p$ and each latent factor has exactly $p/m$ children.
W.l.o.g. we connect the first $p/m$ observed variables with $Z_1$, then next $p/m$ variables with $Z_2$ and so on.
We assume additive white Gaussian noise channel with signal-to-noise ratio $\snr$ from each parent to its children.
In this setup, we set $X_i = \sqrt{\frac{\snr}{\snr+1}} Z_{\pi_i} + \sqrt{\frac{1}{\snr+1}} \eta_i$, where $\pi_i$ is the index of the parent of $X_i$, and $\eta_i$ is independent standard Gaussian noise.
Fig.~\ref{fig:cov_pure_modular} shows a covariance matrix corresponding to a modular latent factor models created using the described procedure.

To create approximately modular latent factor models we do two modifications on a modular latent factor model: correlating the latent variables and adding extra parents for observed variables.
For correlating the latent factors we take $m$ independent standard normal random variables $\xi_j, j=1..m$ and compute $z_j = (\sqrt{2}\xi_j + \xi_u + \xi_v)/2$, where $u, v \sim \text{Uniform}\{1,2,\ldots,m\}$.
For adding extra parents, we randomly sample $p$ extra edges from a latent factor to a non-child observed variable. 
By this we create on average one extra edge per each observed variable.
To keep the notion of clusters well-defined, we make sure that each observed variable has higher mutual information with its main parent compared to that with added extra parents.
Suppose some $X_i$ has $k$ extra parents, $Z_{\tau_1}, \ldots, Z_{\tau_k}$. Then we splits $\frac{s}{s+1}$ -- the variance of the signal in a pure modular case -- into $k+2$ equal parts, $\delta = \frac{s}{(s+1)(k+2)}$.
We then set $X_i = \sqrt{2\delta} Z_{\pi_i} + \sqrt{\delta}Z_{\tau_1} + \cdots + \sqrt{\delta} Z_{\tau_k} + \sqrt{\frac{1}{s+1}} \eta_i$, where again $\eta_i$ is independent standard Gaussian noise.
Figs.~\ref{fig:cov_extraparents},~\ref{fig:cov_correlatedzs},~\ref{fig:cov_extraparents_correlatedzs} show covariance matrices corresponding to approximately modular latent factor models created using the described procedures.

\section{Additional results}
\label{sec:moreresults}

In this section we provide additional results that were not presented in the main text due to the space constraints.

\subsection{Examining the modularity of learned models}
\label{sec:modularitychecking}
\begin{figure}[t]
    \centering
    \begin{subfigure}{0.49\textwidth}
    \includegraphics[width=\textwidth]{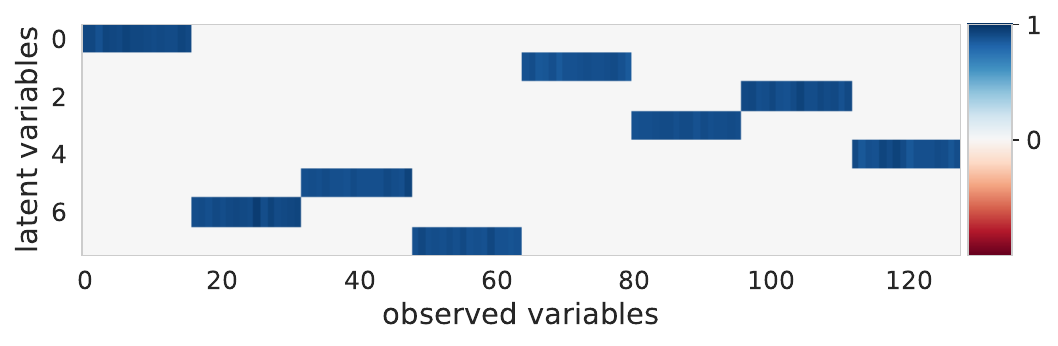}
    \caption{Modular latent factor model}
    \label{fig:mi_pure_modular}
    \end{subfigure}~
    \begin{subfigure}{0.49\textwidth}
    \includegraphics[width=\textwidth]{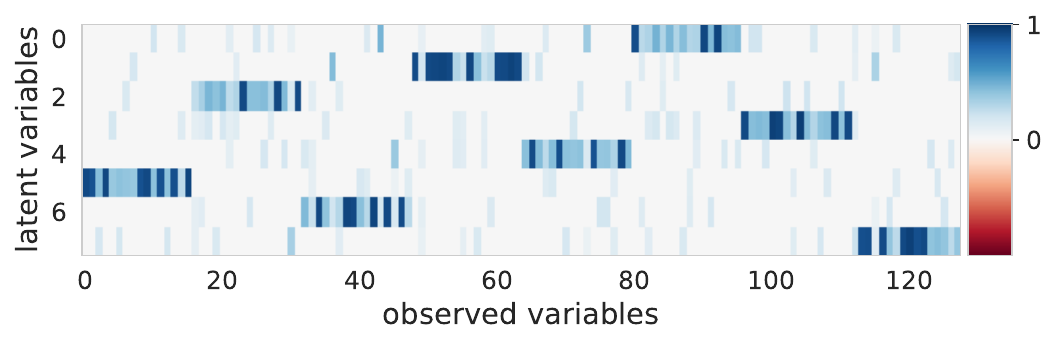}
    \caption{Modular + extra parents}
    \label{fig:mi_extraparents}
    \end{subfigure}
    ~
    \begin{subfigure}{0.49\textwidth}
    \includegraphics[width=\textwidth]{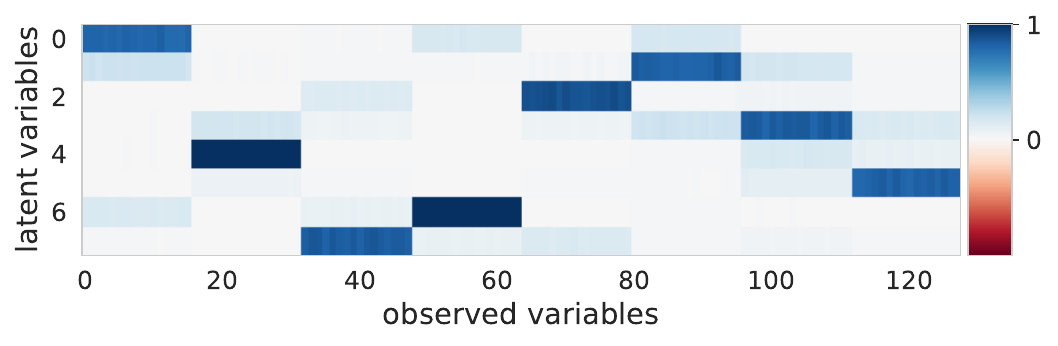}
    \caption{Modular + correlated $Z$s}
    \label{fig:mi_correlatedzs}
    \end{subfigure}~
    \begin{subfigure}{0.49\textwidth}
    \includegraphics[width=\textwidth]{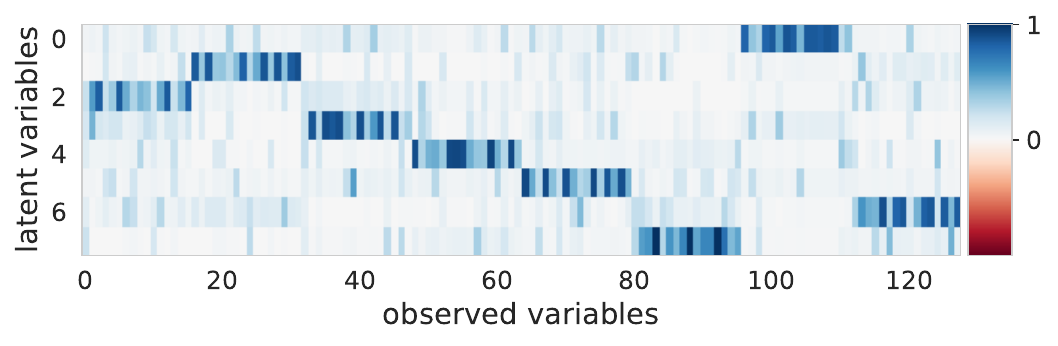}
    \caption{Modular + extra parents and correlated $Z$s}
    \label{fig:mi_extraparents_correlatedzs}
    \end{subfigure}
    \caption{Mutual information matrices between observed variables and latent factors linear CorEx produces when it is trained on a modular (a) and approximately modular (b, c, d) latent factor models. In all examples $m=8$, $p=128$, $\snr=5.0$.} 
    \label{fig:example_mi_matrices}
\end{figure}
We do visualizations to see whether the regularization term of linear CorEx actually leads to learning modular (or approximately modular) latent factor models.
We examine the mutual information matrices between observed and latent variables that linear CorEx produces when it is trained on different types of synthetic data (see Fig~\ref{fig:example_mi_matrices}).
We see that the regularization term we add for encouraging modular structures is indeed effective.

\begin{figure}[t]
    \centering
    \includegraphics[width=\textwidth]{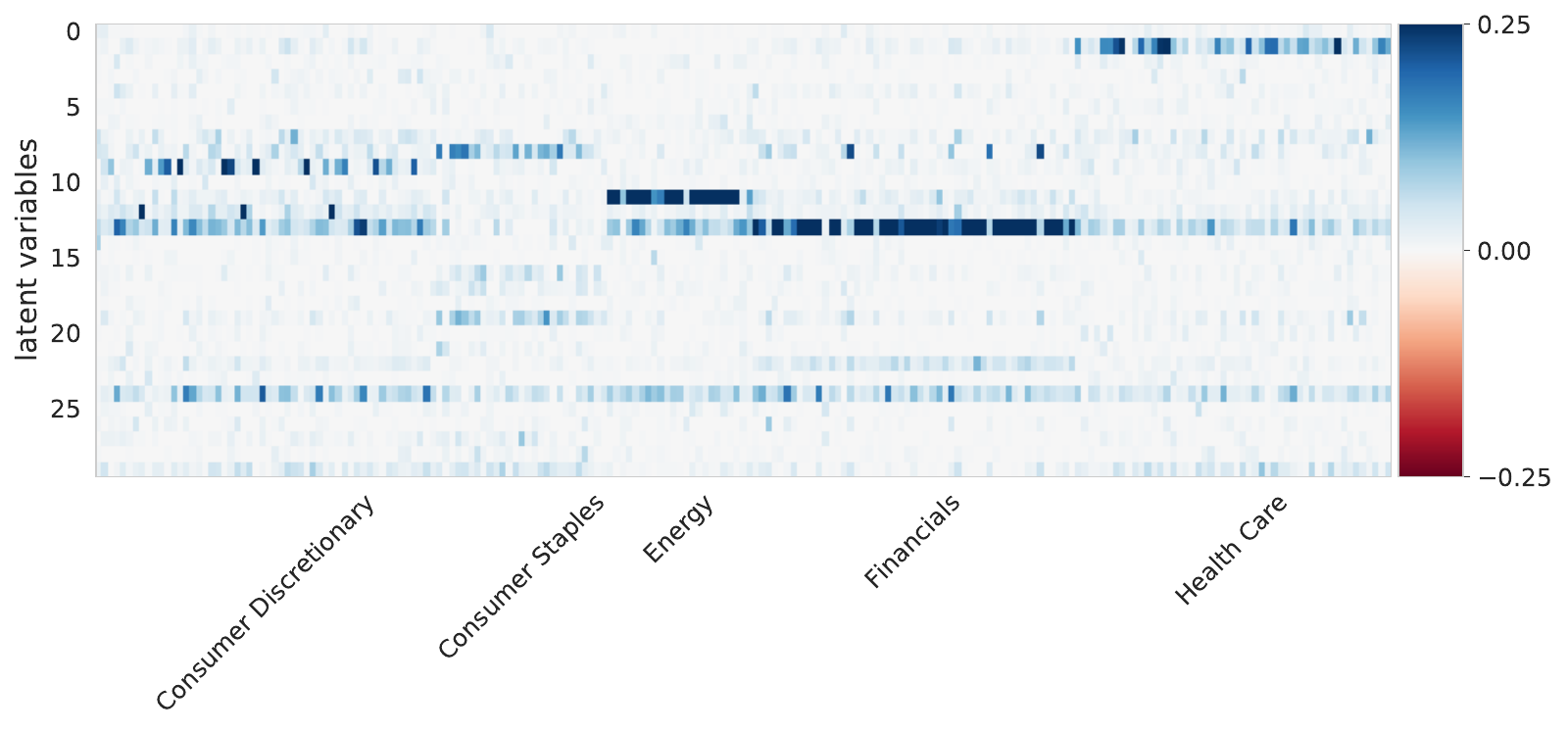}
    \caption{Mutual information matrix between observed variables (stocks) and latent factors linear CorEx produces when trained on the stock marked data (January 2014-January 2017).} 
    \label{fig:mi_stock}
\end{figure}

\begin{figure}[t]
    \centering
    \includegraphics[width=0.7\textwidth]{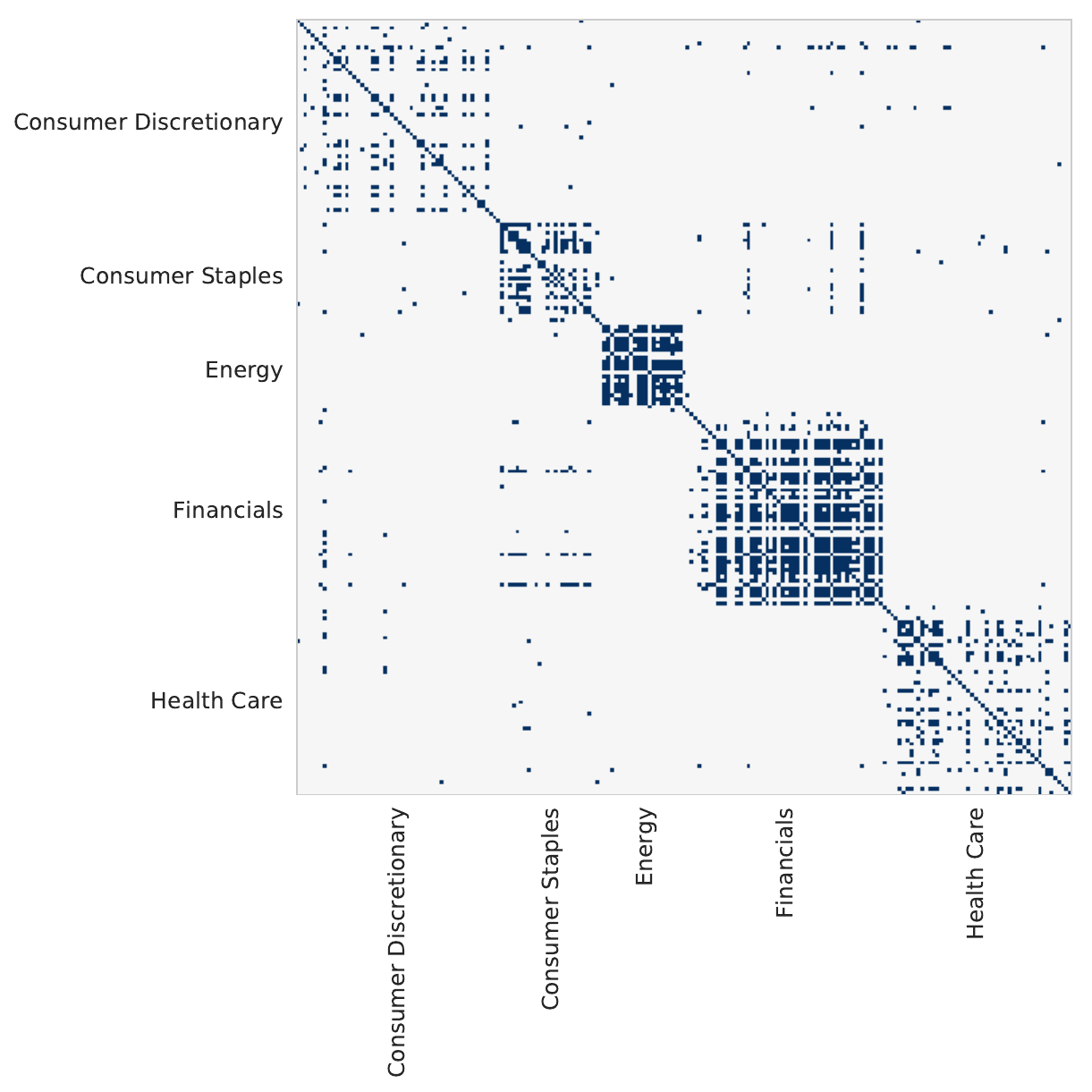}
    \caption{Inverse covariance matrix of some of the S\&P 500 stocks. Plotted are the cells that have absolute value greater than $0.015$.}
    \label{fig:stock_inv_cov}
\end{figure}

Next, we look at the same mutual information matrix for stock 
market data.
Fig.~\ref{fig:mi_stock} shows the mutual information matrix for S\&P 500 stocks belonging to ``consumer discretionary'', ``consumer staples'', ``energy'', ``financials'', and ``health care'' sectors.
We see that most of the stocks have significant mutual information only with a few latent factors.
Moreover, stocks belonging to the same sector are likely to share a parent.
Additionally, we visualize the inverse covariance matrix of these stocks (see Fig.~\ref{fig:stock_inv_cov}).
For Gaussian random variables the thresholded inverse covariance matrix can be interpreted as a random Markov field.
We see that it is almost block-diagonal, but has some off-diagonal connections, confirming that the learned model is close to being a modular latent factor model.

Summing up, all these visualizations assert that the linear CorEx succeeds in biasing the model selection procedure towards modular structures.
More importantly, we see that when the pure modular structure is inappropriate, it picks solutions that are close to being modular.

\subsection{Results on OpenML datasets}
\label{sec:openml_complete}

Table~\ref{tab:results} presents a comparison of various covariance estimation baselines on 51 OpenML datasets.
\begin{table}[t!]
\caption{This table compares covariance estimates on OpenML data. Scores reported are negative log-likelihood (lower better) and the best entry is bolded.
Scores orders of magnitude larger than the best score or evaluating to NaN are shortened with ``*''.
Methods compared, in order, are PCA, Sparse PCA, factor analysis, Ledoit-Wolf, GLASSO (using the BigQUIC algorithm), and the method proposed in this paper.}\label{tab:results}
\centering
\scalebox{0.90}{
\begin{tabular}{@{}lllllllll@{}} \toprule
ID:Dataset & p & n & \multicolumn{6}{c}{Methods} \\ \cmidrule{4-9}
 &  & & PCA& SPCA & FA & LW & BigQUIC & Proposed \\ \midrule
5:arrhythmia & 206 & 54 &178&-33& * &164& * &\bf -74 \\
407:krystek & 1143 & 24 &2122&-1748& * &707&-1428&\bf -2816 \\
408:depreux & 1143 & 20 &1454& * & * &852& * &\bf -482 \\
409:pdgfr & 321 & 63 &112&40& * &83&364&\bf -6 \\
410:carbolenes & 1143 & 29 &1900& * & * &923& * &\bf -8 \\
419:PHENETYL1 & 629 & 17 &876&560&5584&\bf 281&1041&286 \\
420:cristalli & 1143 & 25 &1846& * & * &1366& * &\bf 780 \\
424:pah & 113 & 64 &-129&47&\bf -188&25&134&58 \\
439:chang & 1143 & 27 &2331& * & * &1058& * &\bf -6 \\
1017:arrhythmia & 206 & 54 &178&-33& * &164& * &\bf -60 \\
1104:leukemia & 7129 & 57 &17028&13164&396636&\bf 7019&11530&7336 \\
1107:tumorsC & 7129 & 48 &16990&8499&9642&\bf 8070&9398&8399 \\
1122:APBreastProstat & 10935 & 330 &18427&17219&17741&13431&17002&\bf 10639 \\
1123:APEndometriumBr & 10935 & 324 &18960&12616&12720&11356&18330&\bf 10452 \\
1124:APOmentumUterus & 10935 & 160 &84928&82496&82656&66784&76832&\bf 66176 \\
1125:APOmentumProsta & 10935 & 116 &90024&100392&100032&69168&81264&\bf 67560 \\
1126:APColonLung & 10935 & 329 &84612&76362&76626&\bf 66198&76098&67188 \\
1127:APBreastOmentum & 10935 & 336 &86020&88196&88604&\bf 66824&79968&68408 \\
1128:OVABreast & 10935 & 1236 &83626&79434&79087&\bf 64951&76483&70308 \\
1129:APUterusKidney & 10935 & 307 &85498&74276&74214&\bf 68882&75764&\bf 68882 \\
1130:OVALung & 10935 & 1236 &81989& * &73904&81518&76409&\bf 69291 \\
1131:APProstateUteru & 10935 & 154 &85653&73687&73625&\bf 67208&74617&67363 \\
1132:APOmentumLung & 10935 & 162 &88803&83820&83424&70917&79002&\bf 68805 \\
1133:APEndometriumCo & 10935 & 277 &84840&75376&75152&\bf 65576&76776&66920 \\
1134:OVAKidney & 10935 & 1236 &81964&75144&73507&81592&76210&\bf 69242 \\
1135:APColonProstate & 10935 & 284 &84702&82365&82194&\bf 65550&78318&67260 \\
1136:APLungUterus & 10935 & 200 &87200&77880&77360&69200&76480&\bf 68440 \\
1137:APColonKidney & 10935 & 436 &83776&73515&73084&\bf 66616&75882&68094 \\
1138:OVAUterus & 10935 & 1236 &81964&74772&73358&81493&76136&\bf 69242 \\
1139:OVAOmentum & 10935 & 1236 &81964&75442&74152&81567&76458&\bf 69266 \\
1140:APOvaryLung & 10935 & 259 &87464& * &85280&86320&79560&\bf 68380 \\
1141:APEndometriumPr & 10935 & 104 &90006&85449&84063&69447&77994&\bf 67263 \\
1142:OVAEndometrium & 10935 & 1236 &81989&74574&73086&81493&76062&\bf 69242 \\
1143:APColonOmentum & 10935 & 290 &84332&77198&77140&\bf 65250&76908&66816 \\
1144:APProstateKidne & 10935 & 263 &85277&79553&79553&69589&77857&\bf 68052 \\
1145:APBreastColon & 10935 & 504 &84416&77346&77305&\bf 65165&78144&68448 \\
1146:OVAProstate & 10935 & 1236 &81989&75318&73160&81542&76434&\bf 69266 \\
1147:APOmentumKidney & 10935 & 269 &84780&75114&74952&68796&77112&\bf 67770 \\
1148:APBreastUterus & 10935 & 374 &85500&73275&73155&\bf 66892&79575&68475 \\
1149:APOvaryKidney & 10935 & 366 &86062& * &76590&85470&77626&\bf 68835 \\
1150:APBreastLung & 10935 & 376 &86868&86032&86260&\bf 68195&80332&69441 \\
1151:APEndometriumOm & 10935 & 110 &89474&98868&100386&68178&80388&\bf 66550 \\
1152:APProstateOvary & 10935 & 213 &87118& * &75594&86473&78045&\bf 68026 \\
1153:APColonOvary & 10935 & 387 &85488& * &91572&85254&81198&\bf 68047 \\
1154:APEndometriumLu & 10935 & 149 &89490&74940&75180&70350&76710&\bf 67260 \\
1233:eating & 6373 & 756 &7843&6703&5381&\bf -1110&7980&5457 \\
1457:amazon-commerce & 10000 & 1200 &15576&11376&11256&\bf 10680&12216&10920 \\
1458:arcene & 10000 & 160 &19181&9152&8746&\bf -1267& * &8179 \\
1484:lsvt & 310 & 100 &200&\bf 152&872&212&464&180 \\
1514:micro-mass & 1300 & 288 &1166&50547&50912&1056& * &\bf -708 \\
1515:micro-mass & 1300 & 456 &1260&8041&71493&1224& * &\bf 589 \\
\midrule
\multicolumn{3}{c}{Total \# wins}  &0&1&1&18&0& \bf 32 \\
\bottomrule
\end{tabular}
}
\end{table}

\end{document}